\documentclass[preprint,12pt,authoryear]{elsarticle}

\usepackage{amssymb}
\usepackage{amsmath}
\usepackage{amsmath,amssymb}
\usepackage{graphicx}
\usepackage{booktabs}
\usepackage{multirow}
\usepackage{array}
\usepackage{makecell}
\usepackage{float}
\usepackage{microtype}
\usepackage{adjustbox}
\usepackage{placeins}
\usepackage{tikz}
\usetikzlibrary{arrows.meta,positioning,shadows.blur}

\newcommand{\cyes}{$\checkmark$}
\newcommand{\cno}{$\times$}
\newcommand{\cpart}{$\sim$}
\newcommand{\crecognized}{Recognized}
\newcommand{\cmeasured}{Measured}
\newcommand{\ccontrolled}{Controlled}

\journal{Nuclear Physics B}

\begin{document}

\begin{frontmatter}



\title{Feasible and Novel Synthetic Population Generation with Tabular and Sequential Travel Attributes} 


\author[inst1]{Farbod Abbasi\corref{cor1}}
\ead{farbod.abbasi@mail.concordia.ca}

\author[inst1]{Zachary Patterson}
\ead{Zachary.Patterson@concordia.ca}

\author[inst2]{Bilal Farooq}
\ead{bilal.farooq@torontomu.ca}

\cortext[cor1]{Corresponding author}

\affiliation[inst1]{organization={ Concordia University},
            city={Montreal},
            state={Quebec},
            country={Canada}}

\affiliation[inst2]{organization={Toronto Metropolitan University},
            city={Toronto},
            state={Ontario},
            country={Canada}}

\begin{abstract}
Synthetic populations are critical inputs for activity-based travel demand models, yet generating realistic populations from limited survey data remains challenging. Small samples miss valid attribute combinations, known as sampling zeros, and generative models may also produce infeasible structural zeros. Moreover, realistic synthetic populations must capture both static socio-demographic attributes and sequential travel behaviour, such as trip chains. This paper proposes a regularized two-stage generative framework to address these challenges, where regularization refers to additional loss terms that guide the generator toward broader valid coverage and fewer infeasible samples. In Stage 1, a Wasserstein GAN with gradient penalty is augmented with three regularization terms, IGP, LDR, and CLAP, to improve feasibility, diversity, and novelty in tabular population synthesis. In Stage 2, Transformer and LSTM-Attention models generate sequential travel attributes, including departure time, trip purpose, and travel mode, conditioned on the synthesized tabular profiles. We also introduce novelty and count-aware metrics to evaluate whether valid unseen combinations are recovered and generated in realistic proportions. Results show that regularized models outperform the vanilla WGAN-GP across feasibility, diversity, and novelty. Regularization increases feasibility by 2.1 to 3.7 percentage points and novelty by 6.6 to 10.0 percentage points, improving sampling-zero recovery without sacrificing feasibility. The F1 score improves by 6.3 to 8.6 percentage points. For sequential attributes, LSTM-Attention best matches the trip-length distribution, while Transformer achieves higher overall sequential F1, 90.6\% versus 89.1\%. Cross-stage validation confirms strong consistency between generated mobility status and generated trip chains.
\end{abstract}




\end{frontmatter}



\section{Introduction}
\label{sec1}

Transportation planning increasingly relies on activity-based models (ABMs) to simulate individual-level travel behavior by representing daily activity and trip schedules at a disaggregate level \citep{castiglione2015activity, farooq2013simulation}. ABMs capture the complex interdependencies between socio-demographic characteristics and mobility decisions, enabling more accurate forecasts of travel demand in support of transportation planning and decision-making \citep{both2021activity}. A fundamental requirement of ABMs is a synthetic population that is statistically representative of the true population in the modelled region \citep{agriesti2024assignment}. The quality of this synthetic population directly conditions the realism and reliability of any downstream simulation \citep{bigi2024synthetic}. 

However, generating a reliable synthetic population is a fundamentally difficult problem. Due to privacy concerns and the high cost of data collection \cite{arkangil2022deep}, researchers must rely on limited survey samples that typically cover only one to five percent of the actual population \citep{habib2020large}. These samples, such as public use microdata samples (PUMS) or regional household travel surveys (HTS), provide rich disaggregate information on socio-demographic and behavioural attributes. Crucially, capturing the joint distribution of individual attributes from such data, rather than merely reproducing marginal distributions, is essential for behavioural realism \citep{darsel5295092robust, la2025population}. Even when disaggregate data are available, three interrelated challenges limit the quality of synthesized populations. 

The first concerns the achievement of feasibility, diversity, and novelty in tabular attribute synthesis \citep{darsel5295092robust, bigi2024synthetic, eigenschink2023deep}. Feasibility requires that every generated individual maps to a logically possible combination of attributes. Diversity requires that the generator reproduce the full range of valid attribute combinations present in the true population. Novelty requires that the generator recover valid combinations that exist in the true population but are absent from the training sample due to its limited coverage. These three objectives are in natural tension. A model that aggressively pursues diversity and novelty risks generating infeasible combinations, while a model that is overly conservative may collapse onto a subset of common training patterns and fail to recover unseen combinations. Two key concepts capture this challenge. Sampling zeros are valid attribute combinations that exist in the true population but are absent from the training sample due to limited coverage; a model that fails to recover them produces a population that lacks both diversity and novelty. Structural zeros are logically impossible combinations, such as a child holding a driver's license or a young person in full retirement; a model that generates them produces an infeasible and unrealistic population. Addressing feasibility, diversity, and novelty therefore requires explicit regularization terms added to the loss function to recover sampling zeros while suppressing structural zeros.

The second challenge concerns the type and structure of attributes that a synthetic population must represent \citep{badu2022composite}. Real-world individuals are characterized not only by static tabular attributes, such as age, gender, employment status, and vehicle ownership, but also by sequential behavioural attributes, including trip purpose sequences, departure time patterns, and travel mode chains. These sequential attributes are fundamentally different in structure from tabular ones. They carry temporal dependencies, exhibit variable length, and are closely linked to individual socio-demographic characteristics. A synthetic population that captures only tabular attributes is incomplete and cannot serve as a credible input to activity-based modeling.

The third challenge concerns how the quality of synthetic populations is evaluated. Existing evaluation frameworks primarily assess feasibility and diversity \citep{kim2023deep}. While feasibility is a necessary and valid criterion, relying on diversity alone provides an incomplete picture of generative model performance. Diversity does not reveal where a model succeeds or fails. It does not distinguish between reproducing combinations already seen during training and recovering sampling zeros. A model that achieves high diversity by reproducing common training combinations without recovering sampling zeros may appear to perform well while failing to generalize. We therefore introduce novelty, defined as the share of valid combinations absent from the training sample but present in real population, providing a measure of generalization capacity that diversity alone cannot offer. A further limitation of conventional metrics is that they treat all attribute combinations as equally important regardless of frequency, allowing a model that misrepresents population proportions to still score well. We address this by introducing count-aware variants of both diversity and novelty, which assess whether combinations are generated in proportions consistent with the true population. Together, feasibility, diversity, novelty, and their count-aware counterparts contain a more complete basis for evaluating synthetic population quality.

No existing framework addresses these limitations simultaneously. Studies that propose regularization strategies focus exclusively on tabular attribute generation and do not model sequential behaviour \citep{kim2023deep}. Conversely, studies that generate sequential attributes alongside tabular ones do not incorporate explicit mechanisms to address sampling zeros or structural zeros \citep{badu2022composite, arkangil2022deep}. Moreover, existing evaluation frameworks remain limited, assessing feasibility and conventional diversity while providing no means to separately measure novelty or the distributional accuracy of generated combinations. To the best of our knowledge, no existing framework addresses these three challenges jointly.

This paper addresses all three challenges within a unified two-stage generative framework. In the first stage, a Wasserstein Generative Adversarial Network (WGAN) \citep{arjovsky2017wasserstein} with gradient penalty \citep{gulrajani2017improved} is used to synthesize tabular attributes. To improve feasibility, diversity, and novelty together, we introduce and compare three soft regularization mechanisms incorporated into the generator loss. Unlike hard constraints, these terms do not remove all violations by design; instead, they guide the generator toward a better balance between recovering valid unseen combinations and reducing structurally invalid profiles. In the second stage, a Transformer-based model \citep{vaswani2017attention} and an LSTM with attention mechanism are trained to generate sequential behavioural attributes, namely trip purpose, departure time, and travel mode, conditioned on the tabular attributes. These two architectures are compared to assess which better captures the transition structures and temporal dependencies of real behavioural sequences. The framework is evaluated using the 2018 Montreal Origin-Destination survey, which covers four percent of the regional population and provides both rich socio-demographic and detailed travel diary information.

The main contributions of this study are threefold. First, we propose and evaluate three soft regularization terms for GAN-based tabular population synthesis to improve the balance between feasibility, diversity, and novelty. Second, we extend synthetic population generation beyond static tabular attributes by generating conditional sequential travel attributes using Transformer and LSTM-attention models. Third, we propose a comprehensive evaluation framework that assesses the quality of the generated population across both tabular and sequential dimensions. Together, these contributions provide a more complete basis for generating and evaluating synthetic populations for activity-based travel demand models.

The remainder of this paper is organized as follows. Section 2 reviews the relevant literature on population synthesis methods, deep generative models, diversity and feasibility in synthetic data, and sequential attribute generation. Section 3 describes the Montreal Origin-Destination survey and the case study setup. Section 4 presents the proposed methodology, including the regularized GAN framework and the Transformer and LSTM-based sequence generation models. Section 5 reports and discusses the empirical results across all evaluation dimensions. Section 6 concludes the paper with key insights and directions for future research.

\section{Literature Review}
 
We first discuss traditional and deep learning approaches for tabular attribute synthesis, then examine recent efforts to incorporate sequential behavioural features. Finally, we review strategies for improving diversity and feasibility and identify remaining research gaps.

\subsection{From Traditional Methods to Deep Generative Models}
Early population synthesis methods relied on marginal-fitting techniques, most notably iterative proportional fitting (IPF), which adjust sample weights to match known aggregate totals from census or administrative data\citep{zhu2014synthetic,predhumeau2023synthetic}. While computationally efficient and easy to implement, these methods are fundamentally limited to reproducing marginal distributions and cannot capture the joint dependencies between individual attributes that are essential for behavioural realism. To address this limitation, probabilistic simulation methods were introduced, including Markov Chain Monte Carlo (MCMC) and Hidden Markov Model (HMM) based approaches, which approximate joint distributions by drawing samples from conditional probability structures \citep{farooq2013simulation,saadi2016hidden,kukic2024one}. These methods offer greater flexibility and can generate new individuals beyond those observed in the sample, but they rely on strong structural assumptions and their scalability deteriorates as the number of attributes increases. Bayesian network approaches further advanced joint distribution modelling by explicitly learning dependency structures among attributes\cite{sun2015bayesian,rahman2023population}, while hierarchical and multilevel models extended this to capture household and individual level relationships simultaneously\cite{sun2018hierarchical, kukic2024one}. More recently, integrated pipelines have combined synthesis with spatial assignment and dynamic updating to improve practical applicability \cite{fournier2021integrated,horl2021synthetic,kukic2023hybrid}. Despite these advances, traditional methods share a common limitation: none explicitly addresses sampling zeros and structural zeros, and most are constrained to reproducing combinations already present in the training sample.

\subsection{Diversity and Feasibility in Generative Population Synthesis}

Deep generative models, including VAEs, GANs, diffusion models, and LLMs, have emerged as a promising alternative to traditional methods, owing to their ability to learn complex joint distributions and generate novel individuals beyond those observed in the training sample. However, the extent to which these models address sampling zeros and structural zeros varies considerably. To characterize this variation, we organize existing studies into four groups. Studies in G1 ignore both challenges entirely, treating synthesis as a pure distribution-matching problem \citep{yang2025deep, mensah2025robustness}. Studies in G2 recognize the existence of these challenges but propose no formal metrics or mechanisms to address them \citep{aemmer2022generative, borysov2019generate}. Studies in G3 formally measure sampling and structural zeros, establishing useful evaluation frameworks, but they do not introduce any mechanism to control this during training \citep{garrido2020prediction, jutras2024copula, kang2023generating, johnsen2022population, tang2025generating, rastogi2025population}. Only studies in G4 explicitly control and improve both challenges through mechanisms incorporated into the generative process: \cite{kim2023deep} proposed the first regularization framework combining diversity and feasibility objectives, while \cite{lim2025large} leveraged LLM-based temperature scaling for controllable generation. 

However, even within G4, existing evaluation frameworks remain incomplete in two ways. First, they measure coverage of observed attribute combinations but do not distinguish between combinations seen in the training sample and valid unseen combinations, making it difficult to assess generalization. Second, they often treat all combinations equally and do not evaluate whether generated data reflect realistic population proportions. This study addresses these two limitations by introducing novelty and count-aware metrics. Novelty measures the recovery of valid combinations that are absent from the training sample but present in the full population. The count-aware versions of diversity and novelty further evaluate whether these recovered combinations are generated in proportions that are consistent with the real population.

Crucially, all studies across G1 to G4, regardless of their level of engagement with sampling zeros and structural zeros, focus exclusively on tabular attribute generation and do not model sequential behavioral attributes, a limitation that we address in the following section.

\subsection{Sequential Attribute Generation in Synthetic Populations}

There has been limited research to extend synthetic population generation beyond tabular attributes to include sequential behavioral features such as trip purpose, departure time, and travel mode chains. \cite{badu2022composite} introduced the Composite Travel GAN, the first joint generative framework combining a GAN for tabular attributes with a SeqGAN reinforcement learning module for sequential generation, establishing the foundational architecture for this line of work. \cite{arkangil2022deep} proposed a modular pipeline integrating CTGAN for tabular synthesis with an RNN-based model for trip sequence generation, connected through a Hungarian matching algorithm. \cite{agrawal2025generation} focused specifically on spatio-temporal sequence generation using a multi-headed GAN with GRU, jointly modeling location and time without a tabular component. More recently, \cite{lu2026generate} introduced a multimodal conditional framework combining CTGAN, VQ-VAE, and Transformer architectures with contrastive learning to link individual attributes to activity sequences and locations. Despite these advances, existing studies do not explicitly address sampling zeros or structural zeros in the sequential generation process. This study addresses this gap by proposing a unified two-stage framework for tabular and sequential generation. The framework incorporates explicit mechanisms to improve novelty, diversity and feasibility in the tabular stage, while also introducing novelty and count-aware evaluation metrics to assess how well the generated sequential attributes remain realistic and consistent with the tabular population.

Recent studies have made important progress in synthetic population generation, but several gaps remain. First, most tabular synthesis methods either ignore sampling zeros and structural zeros or only evaluate them after generation, without incorporating mechanisms to control them during training. Second, studies that do address these issues focus mainly on static tabular attributes and do not generate sequential travel behaviour. Third, existing sequential generation approaches model trip chains but do not explicitly address sampling zero recovery and structural zero reduction. Finally, current evaluation frameworks remain limited because they do not fully assess novelty and count-aware performance. These gaps motivate the proposed framework, and Table~\ref{tab:litreview} summarizes how the present study differs from the existing literature.

\begin{table}[!htbp]
\centering
\tiny
\setlength{\tabcolsep}{2.5pt}
\renewcommand{\arraystretch}{1.02}
\caption{Summary of related work on synthetic population generation.
$\checkmark$~= addressed,~$\times$~= not addressed,~$\sim$~= partially addressed.
Samp. = Sampling Zeros; Struct. = Structural Zeros; Seq. = Sequential generation;
Eval. = Evaluation completeness (novelty and count-aware metrics).}
\label{tab:litreview}
\resizebox{\textwidth}{!}{%
\begin{tabular}{>{\raggedright\arraybackslash}p{4.5cm}
                >{\raggedright\arraybackslash}p{3.5cm}
                >{\centering\arraybackslash}p{0.8cm}
                >{\centering\arraybackslash}p{1.5cm}
                >{\centering\arraybackslash}p{1.5cm}
                >{\centering\arraybackslash}p{0.8cm}
                >{\centering\arraybackslash}p{1.2cm}}
\toprule
\textbf{Study} &
\textbf{Model} &
\textbf{Seq.} &
\textbf{Samp. Zeros} &
\textbf{Struct. Zeros} &
\textbf{Grp.} &
\textbf{Eval.} \\
\midrule

\multicolumn{7}{l}{\textbf{Traditional Methods}} \\
\midrule
\cite{zhu2014synthetic}; \cite{predhumeau2023synthetic} & IPF / QISI & \cno & \cno & \cno & Trad. & \cno \\
\cite{farooq2013simulation}; \cite{saadi2016hidden}; \cite{kukic2024one} & MCMC / HMM & \cno & \cno & \cno & Trad. & \cno \\
\cite{sun2015bayesian}; \cite{rahman2023population} & Bayesian Network & \cno & \cno & \cno & Trad. & \cno \\
\cite{sun2018hierarchical}; \cite{kukic2024one} & Hierarchical / Multilevel & \cno & \cno & \cno & Trad. & \cno \\
\cite{horl2021synthetic}; \cite{fournier2021integrated} & Integrated Pipeline & \cno & \cno & \cno & Trad. & \cno \\

\addlinespace[0.25em]
\multicolumn{7}{l}{\textbf{G1: Ignore Sampling \& Structural Zeros}} \\
\midrule
\cite{mensah2025robustness} & CTGAN / VAE & \cno & \cno & \cno & G1 & \cno \\
\cite{yang2025deep} & GAN + DAG & \cno & \cno & \cno & G1 & \cno \\

\addlinespace[0.25em]
\multicolumn{7}{l}{\textbf{G2: Recognize (No Metrics, No Control)}} \\
\midrule
\cite{borysov2019generate} & VAE & \cno & \crecognized & \crecognized & G2 & \cno \\
\cite{aemmer2022generative} & VAE + CVAE & \cno & \crecognized & \cno & G2 & \cno \\

\addlinespace[0.25em]
\multicolumn{7}{l}{\textbf{G3: Measure (No Training-Time Control)}} \\
\midrule
\cite{garrido2020prediction} & WGAN / VAE & \cno & \cmeasured & \cmeasured & G3 & \cno \\
\cite{jutras2024copula} & Copula + GAN & \cno & \cmeasured & \cmeasured & G3 & \cno \\
\cite{kang2023generating} & Diffusion & \cno & \cmeasured & \cpart & G3 & \cno \\
\cite{tang2025generating} & Diffusion & \cno & \cmeasured & \cmeasured & G3 & \cno \\
\cite{rastogi2025population} & CT-GAN & \cno & \cmeasured & \cpart & G3 & \cno \\
\cite{johnsen2022population} & CVAE / CGAN & \cno & \cmeasured & \cno & G3 & \cno \\

\addlinespace[0.25em]
\multicolumn{7}{l}{\textbf{G4: Control \& Improve}} \\
\midrule
\cite{kim2023deep} & Reg. GAN / VAE & \cno & \ccontrolled & \ccontrolled & G4 & \cno \\
\cite{lim2025large} & LLM + BN & \cno & \ccontrolled & \ccontrolled & G4 & \cno \\

\addlinespace[0.25em]
\multicolumn{7}{l}{\textbf{Sequential Generation Studies}} \\
\midrule
\cite{badu2022composite} & GAN + SeqGAN & \cyes & \cno & \cno & Seq. & \cno \\
\cite{arkangil2022deep} & CTGAN + RNN & \cyes & \cno & \cno & Seq. & \cno \\
\cite{agrawal2025generation} & GAN + GRU & \cyes & \cno & \cno & Seq. & \cno \\
\cite{lu2026generate} & CTGAN + VQ-VAE + Transformer & \cyes & \cno & \cno & Seq. & \cno \\

\addlinespace[0.25em]
\multicolumn{7}{l}{\textbf{Present Study}} \\
\midrule
\textbf{Present Study} &
\textbf{Reg. GAN + Transformer / LSTM} &
\cyes &
\ccontrolled &
\ccontrolled &
\textbf{Ours} &
\cyes \\
\bottomrule
\end{tabular}%
}
\end{table}

\section{Data and Case study}
Since access to data on an entire population is not feasible, to perform population synthesis we use the 2018 Montreal Origin Destination (OD) survey, which draws a sample of approximately 4\% of the total population. The OD dataset encompasses detailed travel information for 162,588 individuals. Table \ref{tab:variables} provides descriptive statistics of the OD dataset. It contains 7 individual attributes with a total of 35 categorical classes. 

\begin{table}[H]
\centering
\caption{Tabular socio-demographic attributes and category proportions in the 2018 Montreal OD dataset.}
\label{tab:variables}
\resizebox{0.75\textwidth}{!}{%
\begin{tabular}{l l c}
\hline
\textbf{Attribute (Dimensions)} & \textbf{Category} & \textbf{Proportion (\%)} \\
\hline

\multirow[t]{4}{*}{1. Number of vehicles (m\_auto)}
& 0 & 9.49 \\
& 1 & 36.41 \\
& 2 & 39.95 \\
& 3+ & 14.15 \\[3pt]

\multirow[t]{5}{*}{2. Household size (m\_pers)}
& 1 & 13.03 \\
& 2 & 34.92 \\
& 3 & 18.14 \\
& 4 & 24.03 \\
& 5 & 9.88 \\[3pt]

\multirow[t]{2}{*}{3. Gender (p\_sexe)}
& Male & 48.77 \\
& Female & 51.23 \\[3pt]

\multirow[t]{11}{*}{4. Age group (p\_grage)}
& 0--4 years & 3.63 \\
& 5--9 years & 5.07 \\
& 10--14 years & 5.48 \\
& 15--19 years & 5.12 \\
& 20--24 years & 4.60 \\
& 25--34 years & 8.71 \\
& 35--44 years & 12.66 \\
& 45--54 years & 14.39 \\
& 55--64 years & 18.13 \\
& 65--74 years & 14.03 \\
& 75 years and older & 8.19 \\[3pt]

\multirow[t]{7}{*}{5. Employment status (p\_statut)}
& Full-time worker & 40.71 \\
& Part-time worker & 4.67 \\
& Student & 19.55 \\
& Retired & 26.43 \\
& Other & 2.74 \\
& children under 4 years & 3.63 \\
& At home & 2.27 \\[3pt]

\multirow[t]{3}{*}{6. Driver's license (p\_permis)}
& Yes & 72.52 \\
& No & 12.30 \\
& Not applicable & 15.19 \\[3pt]

\multirow[t]{3}{*}{7. Mobility (p\_mobil)}
& Yes & 78.71 \\
& No & 17.66 \\
& children under 4 years & 3.63 \\

\hline
\end{tabular}%
}
\end{table}

\begin{figure}[htbp]
    \centering
    \includegraphics[width=\textwidth]{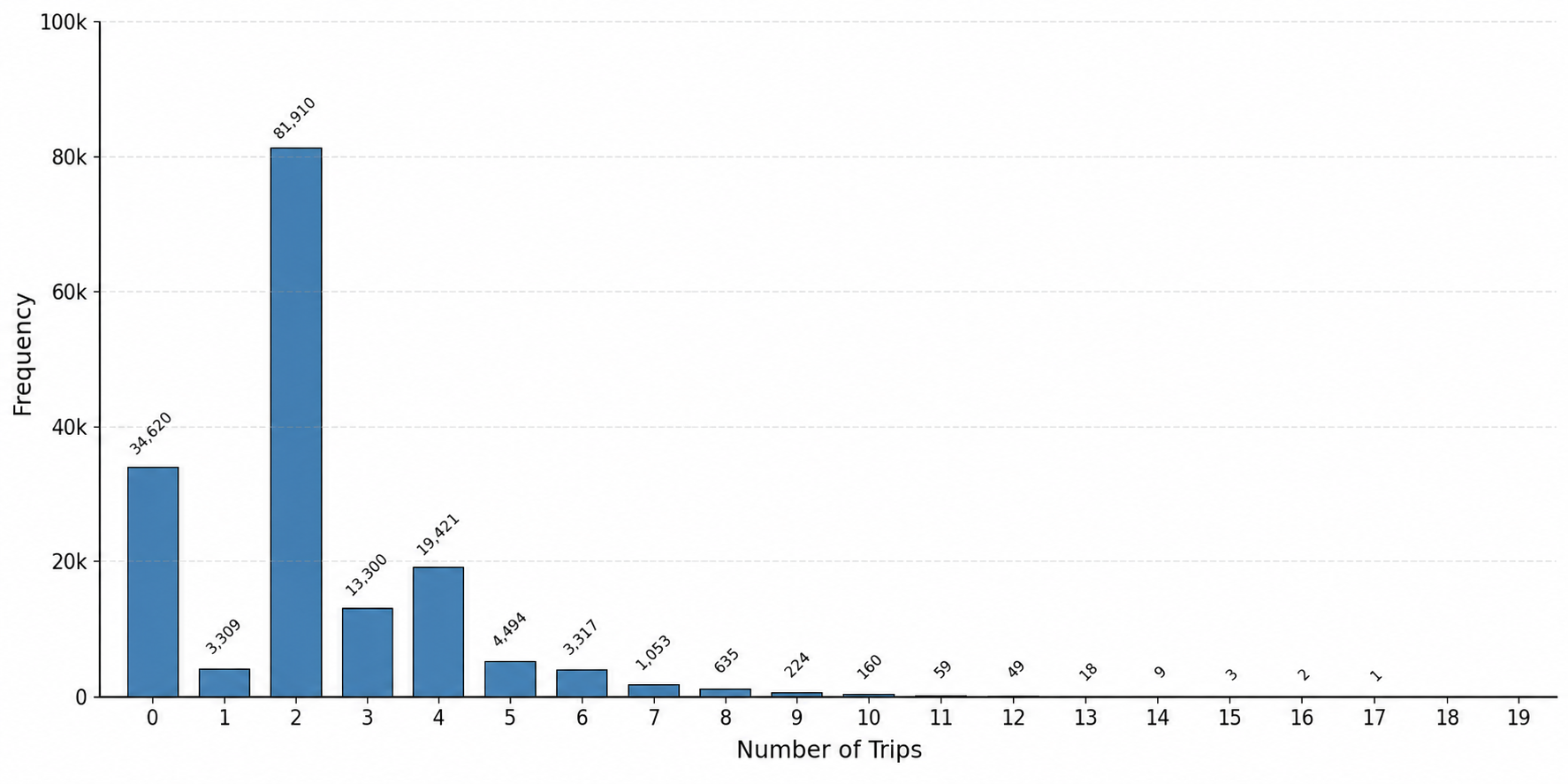}
    \caption{Trip Sequence Length Distribution}
    \label{fig:trip length}
\end{figure}

In addition to these static attributes, the dataset contains three sequential trip-level features: departure time, trip purpose, and trip mode. These sequential variables represent each individual’s daily travel behaviour as an ordered trip chain. The sequence length represents the total number of trips made by an individual. Figure \ref{fig:trip length} illustrates the distribution of trip sequence lengths in the OD dataset. As shown, the majority of individuals make between zero and four trips per day, while longer trip chains occur less frequently.

From this point onward, the OD dataset is treated as a representative sample of the full population. A subset of this dataset is designated as training data, and our objective is to generate synthetic samples that can replicate the complete set of individuals present in the full OD dataset. We can reasonably assume that if we are able to accurately reconstruct the full OD dataset using only a subset of it as training data, then it would also be feasible to generate the entire population through the OD dataset itself. 

The concept of sampling zeros refers to valid feature combinations that exist in the full population but are missing from the training dataset due to limited sample size. These are not infeasible combinations, and they simply do not appear in the sample, which poses a challenge for generative models attempting to replicate the full population accurately.

\begin{figure}[H]
    \centering
    \includegraphics[width=0.8\textwidth]{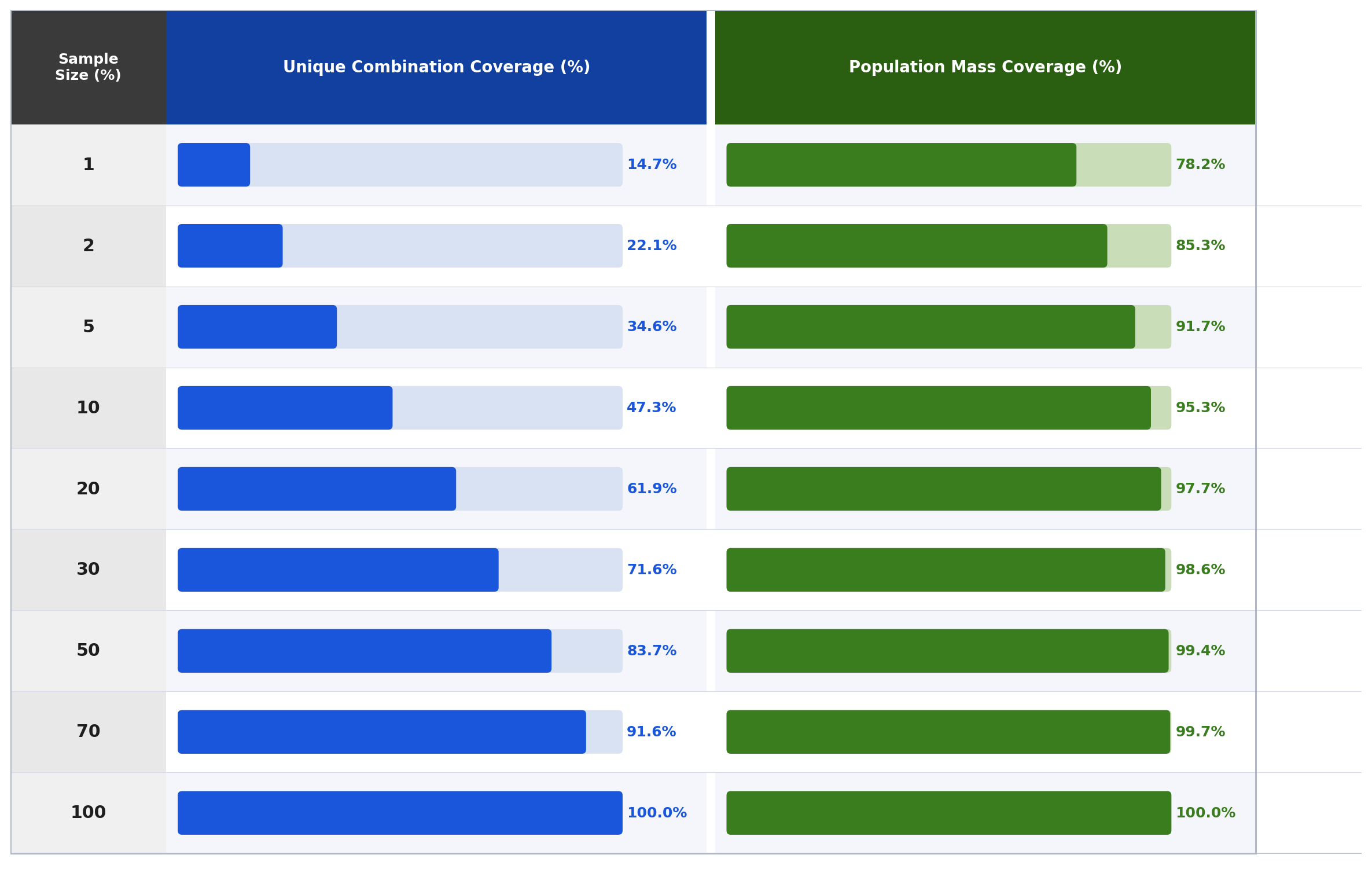} 
    \caption{Relationship between sample size and sampling zeros}
    \label{fig:sampling_zeros}
\end{figure}

Figure ~\ref{fig:sampling_zeros} illustrates this issue. The blue bars represent unique combination coverage, which is the percentage of unique feature combinations in the full dataset captured at each sample size. As the sample size increases, a larger proportion of the unique combinations present in the full dataset is recovered. Correspondingly, The green bars represent population mass coverage, defined as the proportion of total individuals in the OD dataset who fall into these captured combinations. For example, at a 1\% sample size, 14.7\% of the unique feature combinations are observed. However, these combinations account for 78.2\% of the individuals in the dataset. This highlights that the training data lacks diversity, and therefore, a synthetic model trained on it must be capable of generating plausible combinations not seen in the training samples.

We focus our analysis on the 1\% sample size. To claim that a synthetic population is truly reliable, it must increase the number of unique feature combinations beyond what is observed in the training data and recover the 11.8\% of individual records from the full OD dataset that are not included in the training set, by generating samples that match with missing combinations.

\section{Methodology}

\subsection{Framework Overview}

This paper proposes a two-stage generative framework for synthesizing individual-level travel behavior profiles consistent with the distributional properties of an observed OD survey. Let each individual in the population be represented by a pair $(X, S)$, where $X \in \mathbb{R}^d$ denotes a vector of tabular attributes and $S = (s_1, s_2, \dots, s_T)$ denotes a sequence of length $T$. The objective of synthetic population generation is to learn the joint distribution $P(X, S)$ from observed data and generate new synthetic samples $(\tilde{X}, \tilde{S})$ that preserve statistical realism, novelty, and behavioural feasibility. Directly modelling the joint distribution of tabular and sequential attributes is challenging due to their different structures. To address this, we decompose the joint distribution as:

\begin{equation}
P(X, S) = P(X) P(S \mid X).
\end{equation}

\begin{figure}[htbp]
    \centering
    \includegraphics[width=\textwidth]{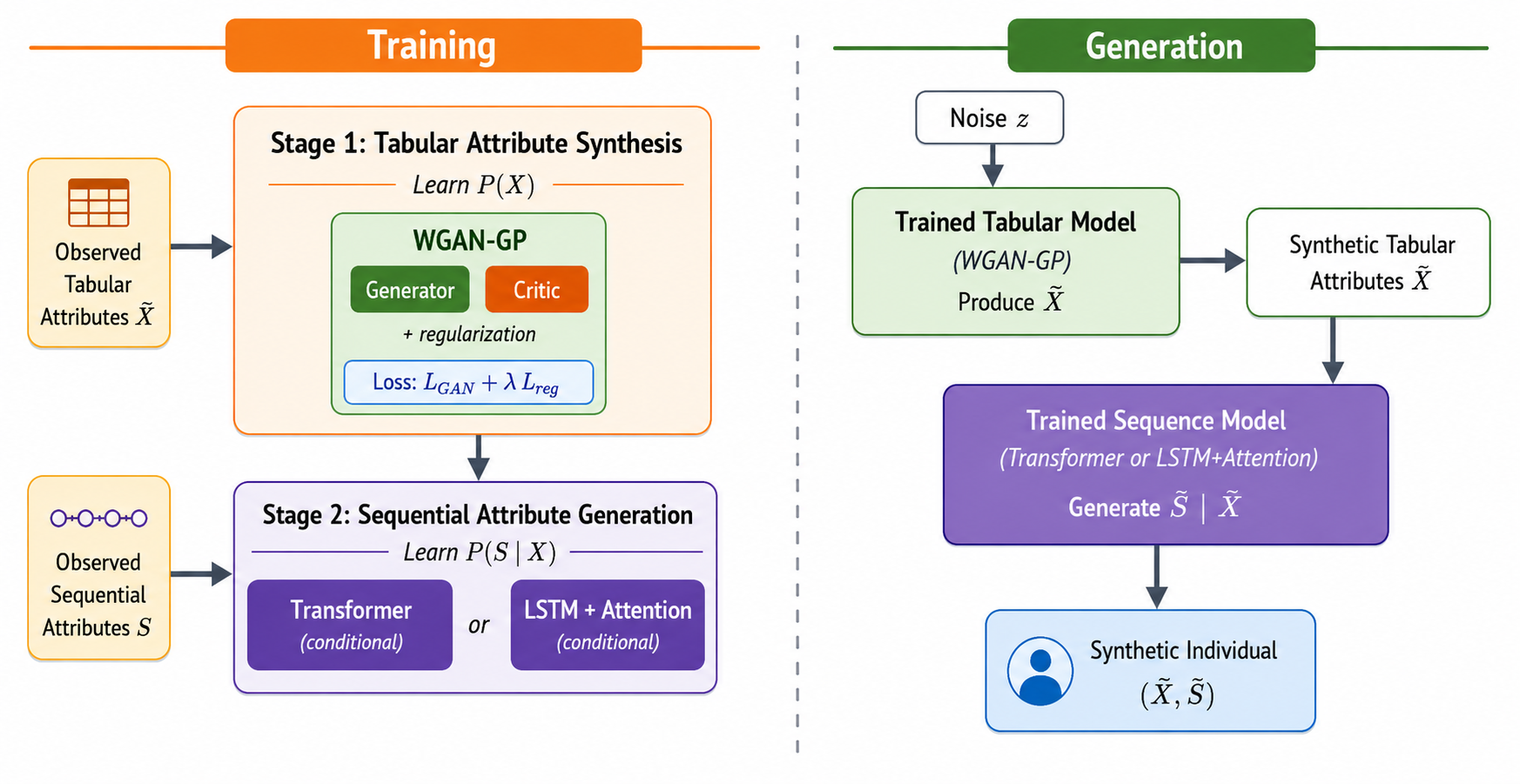}
    \caption{Two stage framework for synthetic population generation}
    \label{fig:diagram}
\end{figure}

Based on Figure \ref{fig:diagram}, two generative models are trained to learn the distribution of tabular attributes. In the first stage, a Wasserstein GAN \citep{arjovsky2017wasserstein} with Gradient Penalty \citep{gulrajani2017improved} (WGAN-GP), enhanced with three regularization terms to improve the coverage of valid attribute combinations. In the second stage, a Transformer model and an LSTM model augmented with Attention are trained to generate sequences attributes conditioned on tabular attributes.

\subsection{Stage 1: Tabular Attribute Synthesis}

\subsubsection{WGAN-GP Architecture}

In the first stage of the proposed framework, a WGAN-GP is used to generate tabular socio-demographic attributes. WGAN-GP is adopted because it provides a more stable training process than the standard GAN \citep{al2024optimization} and reduces the risk of mode collapse by optimizing the Wasserstein distance between the real and generated data distributions \citep{thanh2020catastrophic}.

A GAN consists of two competing neural networks: a generator and a critic. The generator $G$ receives a random noise vector $z \sim p_z(z)$ from the latent space and maps it into a synthetic individual profile $\tilde{x}=G(z)$. In this study, each generated profile represents a combination of categorical tabular attributes such as household size, vehicle ownership, age group, employment status, driver's license status, gender, and mobility status. The critic $D$, instead of classifying samples as real or fake, assigns a scalar score to each sample and learns to distinguish the distribution of real individuals from the distribution of generated individuals.

The WGAN objective encourages the generator to produce synthetic samples that receive high critic scores, while the critic is trained to assign higher scores to real samples than to generated ones. To improve training stability and enforce the Lipschitz constraint required by WGAN, a gradient penalty term is added to the critic loss. The critic loss is defined as:

\begin{equation}
\mathcal{L}_{D}
=
\mathbb{E}_{z \sim p_z(z)}[D(G(z))]
-
\mathbb{E}_{x \sim p_r(x)}[D(x)]
+
\lambda_{GP}
\mathbb{E}_{\hat{x}}
\left[
\left(
\|\nabla_{\hat{x}}D(\hat{x})\|_2 - 1
\right)^2
\right],
\label{eq:critic_loss}
\end{equation}

where $x$ denotes a real sample, $G(z)$ denotes a generated sample, $\hat{x}$ is an interpolated sample between real and generated data, and $\lambda_{GP}$ controls the strength of the gradient penalty.

The standard generator loss in WGAN-GP is defined as:

\begin{equation}
\mathcal{L}_{G}
=
-
\mathbb{E}_{z \sim p_z(z)}[D(G(z))].
\label{eq:generator_loss}
\end{equation}

This adversarial loss encourages the generator to create synthetic samples that are increasingly similar to the real population distribution. However, matching the overall distribution alone does not guarantee that the generated population is sufficiently diverse, feasible, or novel. In particular, a generator may still fail to recover valid combinations that are absent from the training sample, or it may generate unrealistic attribute combinations.

To address this limitation, we extend the generator objective by incorporating an additional regularization term:

\begin{equation}
\mathcal{L}_{G}^{reg}
=
-
\mathbb{E}_{z \sim p_z(z)}[D(G(z))]
+
\lambda_{reg}\mathcal{L}_{reg},
\label{eq:regularized_generator_loss}
\end{equation}

where $\mathcal{L}_{reg}$ denotes one of the proposed regularization terms and $\lambda_{reg}$ controls its contribution to the generator loss. These regularization terms are designed to guide the generator toward better exploration of the latent space, improving the recovery of rare but valid combinations while preserving feasibility.

The following section describes the proposed regularization terms used to enhance diversity, novelty, and feasibility in tabular population synthesis.

\subsubsection{Regularization Terms}

A key challenge in tabular population synthesis is that a small training sample cannot represent all valid combinations that exist in the full population. In this study, the model is trained using only a 1\% sample of the OD dataset. As a result, many valid but low-frequency combinations may be absent from the training data. A standard WGAN-GP may therefore learn to reproduce only the most frequent combinations and fail to generate rare but realistic individuals. This limits the diversity and novelty of the generated population.

\begin{figure}[htbp]
    \centering
    \includegraphics[width=\textwidth]{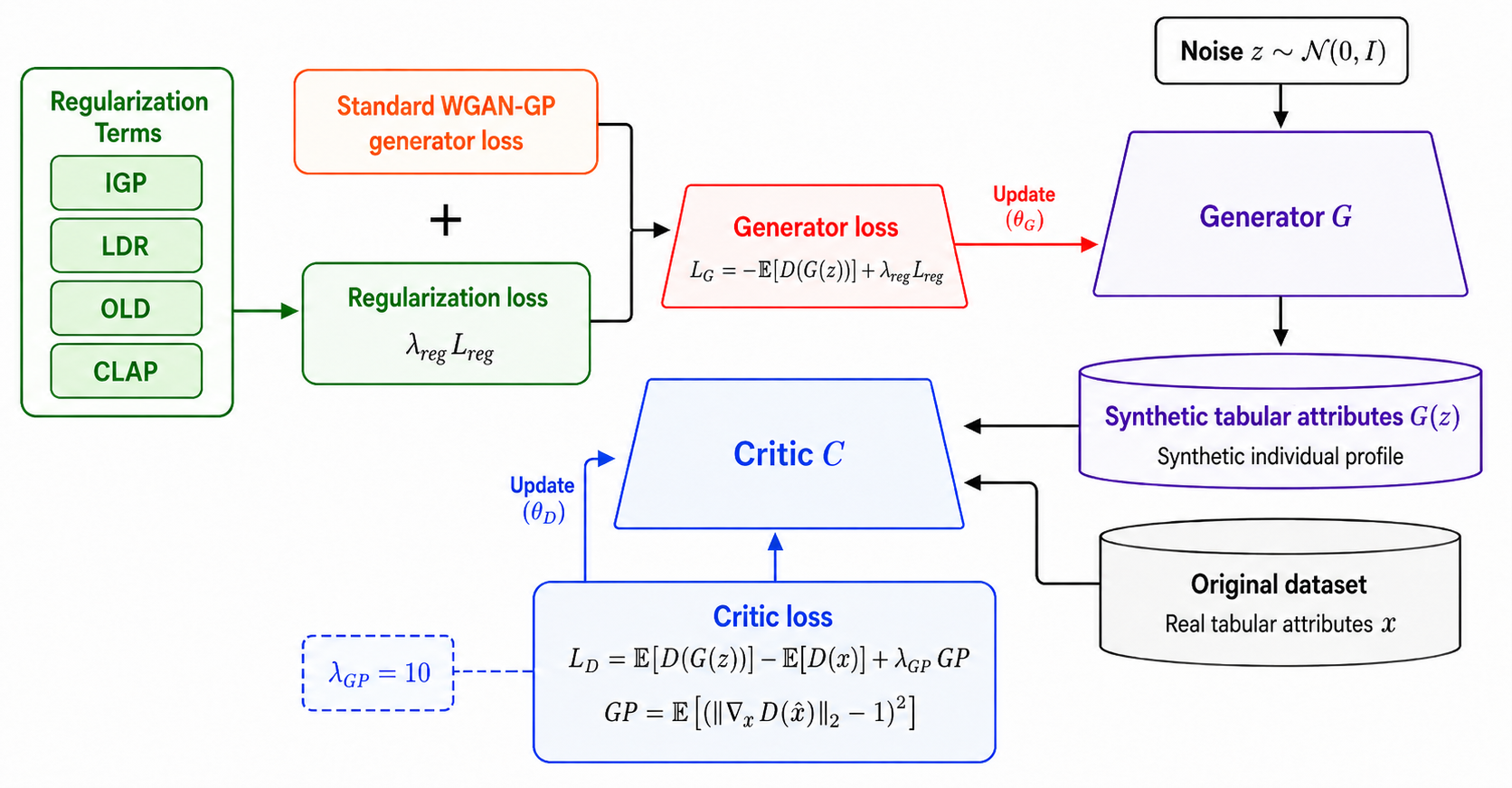}
    \caption{Overview of the proposed regularized WGAN-GP framework for tabular attribute synthesis}
    \label{fig:stage1_wgan}
\end{figure}

To address this issue, three regularization terms are incorporated into the WGAN-GP generator loss introduced in the previous section. These terms are designed to encourage the generator to make better use of the latent space and produce a wider range of valid attribute combinations. Each regularization term is tested separately in order to evaluate its effect on diversity, novelty, and feasibility. Figure~\ref{fig:stage1_wgan} provides a schematic overview of the regularized WGAN-GP used in Stage 1 for tabular attribute synthesis.

Regularization provides a flexible way to guide the generator during training, but it should not be interpreted as a hard feasibility constraint. The proposed terms cannot guarantee that all structurally invalid profiles will be removed or that all missing valid combinations will be recovered. Instead, they act as soft penalties in the generator loss and encourage more useful exploration of the latent space. This is important because the valid population space is only partially observed in the training sample. Therefore, strict constraints may improve feasibility but can also prevent the model from recovering rare but valid combinations.

Compared with existing distance-based regularization approaches \citep{kim2023deep}, the proposed terms place more emphasis on controlled exploration and novelty. Prior work mainly uses regularization to reduce infeasible generation by discouraging samples that move far from valid regions of the data space. In contrast, our approach encourages the generator to explore broader latent-space mappings, to recover valid combinations absent from the limited training sample. Therefore, The goal is to improve novelty and diversity without reducing feasibility.

\paragraph{Inverse Gradient Penalty (IGP)}

The IGP term encourages the generator to be sensitive to changes in the latent space. The motivation is that if two latent vectors are different, their generated outputs should also be sufficiently different. Otherwise, the generator may ignore parts of the latent space and map many different latent inputs to the same or very similar outputs.

For two latent vectors \(z_1\) and \(z_2\), IGP compares the distance between their generated outputs with the distance between the latent vectors themselves. The loss is defined as:

\begin{equation}
\mathcal{L}_{IGP}
=
-
\mathbb{E}_{z_1,z_2}
\left[
\min
\left(
\frac{
\left\|
G(z_1)-G(z_2)
\right\|_2
}{
\left\|
z_1-z_2
\right\|_2
},
\tau
\right)
\right],
\label{eq:igp}
\end{equation}

where \(\tau\) is a threshold that limits the maximum value of the ratio. A larger ratio means that changes in the latent space lead to meaningful changes in the generated output. Because the loss has a negative sign, minimizing it encourages the generator to increase this ratio up to the threshold. In this way, IGP helps the generator use the latent space more effectively and produce a more diverse and novel samples.

\paragraph{Latent Diversity Regularization (LDR)}

The LDR term directly encourages nearby points in the latent space to generate different outputs. Without this regularization, small changes in the latent vector may produce almost identical individuals, which indicates local mode collapse. LDR reduces this problem by rewarding the generator when a small movement in the latent space leads to a noticeable change in the generated output.

For a latent vector \(z\) and a small perturbation \(\delta\), the LDR loss is defined as:

\begin{equation}
\mathcal{L}_{LDR}
=
-
\mathbb{E}_{z,\delta}
\left[
\left\|
G(z+\delta)-G(z)
\right\|_2
\right],
\label{eq:ldr}
\end{equation}

The negative sign means that minimizing this loss encourages the distance between \(G(z+\delta)\) and \(G(z)\) to become larger. In other words, the generator is encouraged to produce more distinct samples for nearby latent inputs. This improves local diversity and helps the model explore more valid combinations.

\paragraph{Cross-Latent Agreement Penalty (CLAP)}

The CLAP term promotes global diversity by discouraging the generator from producing similar outputs for different latent vectors. While LDR focuses on nearby points in the latent space, CLAP compares outputs generated from two independently sampled latent vectors. If two different latent vectors produce very similar synthetic individuals, the model receives a larger penalty.

For two latent vectors \(z_1\) and \(z_2\), the CLAP loss is defined as:

\begin{equation}
\mathcal{L}_{CLAP}
=
\mathbb{E}_{z_1,z_2}
\left[
\exp
\left(
-
\left\|
G(z_1)-G(z_2)
\right\|_2
\right)
\right].
\label{eq:clap}
\end{equation}

This penalty is large when the generated outputs are close to each other and becomes smaller as the outputs become more different. Therefore, CLAP encourages the generator to spread generated samples across the output space and reduces duplication among generated individuals.

Overall, the four regularization terms target diversity and novelty from different perspectives. These regularized WGAN-GP variants are compared with the vanilla WGAN-GP to assess whether they improve the recovery of valid missing combinations while maintaining feasibility. 

\subsection{Stage 2: Sequential Attribute Generation}

Stage 2 generates sequential behavioural attributes conditioned on the tabular profile produced in Stage 1. Each individual is associated with three aligned categorical sequences: departure time group, trip purpose, and main travel mode. Let $X$ denote the tabular attribute vector, and let $S^{\text{time}}$, $S^{\text{purpose}}$, and $S^{\text{mode}}$ denote the three behavioral sequences.

The conditional joint distribution of these aligned sequences is modelled autoregressively as:
\begin{equation}
P(S^{\text{time}}, S^{\text{purpose}}, S^{\text{mode}} \mid X)
=
\prod_{t=1}^{T}
P\!\left(
s_t^{\text{time}}, 
s_t^{\text{purpose}}, 
s_t^{\text{mode}}
\mid
s_{<t}^{\text{time}}, 
s_{<t}^{\text{purpose}}, 
s_{<t}^{\text{mode}}, 
X
\right).
\end{equation}

This formulation reflects that, at each trip index $t$, the model predicts three aligned behavioural attributes rather than a single token. The history available to the model consists of all previously generated departure time, purpose, and mode tokens, together with the tabular attributes.

\subsubsection{Transformer Model}

The Transformer model conditions on the tabular attributes by mapping $X$ into a context representation, which is inserted as a prefix token at the beginning of the sequence. The embedded trip tokens are then appended to this context token and passed through a causally masked Transformer encoder. The causal mask ensures that each trip position can attend only to the tabular context and previous trips, preserving the autoregressive structure of the generation process.

At each trip index $t$, the Transformer produces a shared hidden representation $h_t$. From this shared representation, three separate output heads generate probability distributions over departure time, trip purpose, and travel mode:
\begin{equation}
P\!\left(
s_t^{\text{time}}
\mid
s_{<t}^{\text{time}},
s_{<t}^{\text{purpose}},
s_{<t}^{\text{mode}},
X
\right)
=
\mathrm{Softmax}(W_{\text{time}} h_t),
\end{equation}

\begin{equation}
P\!\left(
s_t^{\text{purpose}}
\mid
s_{<t}^{\text{time}},
s_{<t}^{\text{purpose}},
s_{<t}^{\text{mode}},
X
\right)
=
\mathrm{Softmax}(W_{\text{purpose}} h_t),
\end{equation}

\begin{equation}
P\!\left(
s_t^{\text{mode}}
\mid
s_{<t}^{\text{time}},
s_{<t}^{\text{purpose}},
s_{<t}^{\text{mode}},
X
\right)
=
\mathrm{Softmax}(W_{\text{mode}} h_t),
\end{equation}
where $W_{\text{time}}$, $W_{\text{purpose}}$, and $W_{\text{mode}}$ are head-specific projection matrices corresponding to their respective vocabularies.

Let $\mathcal{K} = \{\text{time}, \text{purpose}, \text{mode}\}$. The negative log-likelihood sequence objective is written as:
\begin{equation}
\mathcal{L}_{seq}
=
-
\sum_{t=1}^{T}
\sum_{k \in \mathcal{K}}
\log
P\!\left(
s_t^{k}
\mid
s_{<t}^{\text{time}},
s_{<t}^{\text{purpose}},
s_{<t}^{\text{mode}},
X
\right).
\end{equation}

This shared-representation and multi-head output structure allows the model to capture dependencies among departure time, trip purpose, and travel mode while preserving the categorical structure of each behavioural attribute. During inference, the Transformer generates the sequence autoregressively, sampling the three trip attributes at each step from their predicted distributions.

\subsubsection{LSTM with Attention Model}

Although the Transformer model provides a flexible architecture for capturing long-range dependencies through self-attention, it is also more complex and computationally demanding. Therefore, we also implement an LSTM with attention as a second sequence-generation model. This model provides a recurrent alternative that is simpler in structure while still allowing the use of attention over previous trips. Comparing the Transformer and LSTM-attention models allows us to evaluate whether the additional complexity of the Transformer leads to better sequential attribute generation performance.

The LSTM-attention model conditions on the tabular attributes by initializing the recurrent hidden and cell states. The embedded trip sequence is then processed by the LSTM:
\begin{equation}
O =
\mathrm{LSTM}
\left(
[e_1,\ldots,e_T]; h_0, c_0
\right),
\end{equation}
where $O=(o_1,\ldots,o_T)$ denotes the sequence of LSTM outputs, and $h_0$ and $c_0$ are initialized from the tabular attributes.

A causal attention layer is applied to the LSTM outputs so that each trip position can selectively use information from previous trips while preventing access to future trips:
\begin{equation}
A =
\mathrm{softmax}
\left(
\frac{OO^{T}}{\sqrt{d_{model}}}+M
\right)O,
\end{equation}
where $M$ is a causal mask that blocks attention to future positions. The attended representation is then refined using residual connections, layer normalization, and a feedforward layer.

The final hidden representation at each trip index is passed through the same three-head output structure used in the Transformer model to predict departure time, trip purpose, and travel mode. During inference, the model generates the sequence one trip at a time while maintaining its recurrent state and applying attention over the generated history.

Overall, the Transformer relies fully on self-attention to model dependencies across the trip chain, while the LSTM-attention model combines recurrent memory with attention-based refinement. Comparing these two models allows us to evaluate which architecture better captures sequential travel behavior conditioned on tabular attributes.

\subsection{Evaluation Framework}

To evaluate the quality of the synthetic population, we use a framework that considers both tabular and sequential attributes. For tabular attributes, we evaluate feasibility, diversity, novelty, and the count-aware variants of diversity and novelty. For sequential attributes, we assess whether the generated trip chains reproduce the temporal patterns observed in the real OD data. 

Let $C_{real}$ denote the set of unique tabular attribute combinations observed in the full OD dataset, $C_{sample}$ the set of combinations observed in the training sample, and $C_{fake}$ the set of combinations observed in the synthetic population. Similarly, let $n_{real}(c)$, $n_{sample}(c)$, and $n_{fake}(c)$ denote the number of individuals with combination $c$ in the full OD dataset, training sample, and synthetic population, respectively. The set of valid combinations that are absent from the training sample is defined as:
\begin{equation}
C_{unseen}=C_{real}-C_{sample}.
\end{equation}
This set represents sampling zero combinations. Combinations that exist in the full population but are not observed in the limited training sample.

\subsubsection{Feasibility}

Feasibility measures whether the generated individuals correspond to valid attribute combinations. In this study, a generated combination is considered feasible if it appears at least once in the full OD dataset. Feasibility is defined as:
\begin{equation}
\mathrm{Feasibility}
=
\frac{\sum_{c\in C_{real}} n_{fake}(c)}{N_{fake}}.
\end{equation}
A value of 1 indicates that all generated individuals belong to combinations observed in the full OD dataset. Lower values indicate that the model has generated invalid or structurally unrealistic tabular profiles.

\subsubsection{Diversity and Count-Aware Diversity}

Diversity evaluates how much of the full population's combinatorial support is recovered by the synthetic population. It is measured as the fraction of real combinations that also appear in the generated data:
\begin{equation}
\mathrm{Diversity}
=
\frac{|C_{real}\cap C_{fake}|}{|C_{real}|}.
\end{equation}
This metric captures whether the model can generate a wide range of valid attribute combinations. However, it only considers whether a combination appears or not, and does not account for how frequently each combination is generated.

To address this limitation, we also use count-aware diversity, which compares the generated and real counts for each valid combination:
\begin{equation}
\mathrm{Diversity}_w
=
\frac{\sum_{c\in C_{real}} \min(n_{real}(c),n_{fake}(c))}
{\sum_{c\in C_{real}} n_{real}(c)}.
\end{equation}
This metric rewards the model not only for recovering valid combinations, but also for generating them in proportions that are consistent with the full OD dataset. Therefore, the count-aware diversity score provides a more informative and stricter evaluation than the standard diversity score, because it accounts not only for whether combinations are recovered, but also for whether they are generated with realistic frequencies.

\subsubsection{Novelty and Count-Aware Novelty}

Novelty measures the ability of the model to recover valid combinations that were not present in the training sample. These combinations are important because they correspond to sampling zeros. Novelty is defined as:
\begin{equation}
\mathrm{Novelty}
=
\frac{|C_{unseen}\cap C_{fake}|}{|C_{unseen}|}.
\end{equation}
A higher novelty score indicates that the model is better able to generalize beyond the observed training combinations and recover valid but unseen profiles.

As with diversity, the standard novelty metric only measures whether unseen combinations are recovered, regardless of their frequency. Therefore, we also compute count-aware novelty:
\begin{equation}
\mathrm{Novelty}_w
=
\frac{\sum_{c\in C_{unseen}} \min(n_{real}(c),n_{fake}(c))}
{\sum_{c\in C_{unseen}} n_{real}(c)}.
\end{equation}
This metric evaluates whether the model recovers unseen valid combinations in proportions that are consistent with their frequency in the full OD dataset. Together, novelty and count-aware novelty provide a direct measure of the model's capacity to recover sampling-zero combinations.

\subsubsection{Sequential Evaluation Metrics}

In addition to tabular quality, the generated population must reproduce realistic trip chain behaviour. We therefore evaluate the sequential attributes from two complementary perspectives.

First, we compare the trip length distribution of the generated data against the real OD dataset. For each individual sequence, the trip length is defined as the number of trips in the sequence. This evaluation measures whether the generated population reproduces the overall number of trips per individual. Let $L_R(k)$ and $L_G(k)$ denote the number of real and generated records with trip length $k$, respectively. The trip length distribution is then compared visually using bar charts across the real, and generated datasets. Furthermore, we assess whether the generated population reproduces complete and interpretable activity-sequence patterns. For this evaluation, we identify the most frequent trip-purpose sequences in the real OD dataset and compare their relative shares with the corresponding shares in the generated population.

To further evaluate the sequential quality of the generated data, we analyze the diversity, novelty, and feasibility of the generated sequence attributes. For each sequential attribute, we extract unigrams, bigrams, and trigrams from the real data, the generated data, and the random sample. Unigrams represent individual sequence elements, bigrams capture pairwise transitions, and trigrams represent longer sequential patterns.

\begin{figure}[H]
\centering
\begin{tikzpicture}[
    node distance=1.0cm and 1.25cm,
    main/.style={
        rectangle,
        rounded corners=7pt,
        draw=none,
        fill=gray!18,
        blur shadow={shadow blur steps=5, shadow xshift=1.2pt, shadow yshift=-1.2pt},
        align=center,
        minimum width=3.3cm,
        minimum height=0.9cm,
        font=\small\bfseries
    },
    input/.style={
        rectangle,
        rounded corners=7pt,
        draw=none,
        fill=blue!12,
        blur shadow={shadow blur steps=5, shadow xshift=1.2pt, shadow yshift=-1.2pt},
        align=center,
        minimum width=3.0cm,
        minimum height=0.85cm,
        font=\small
    },
    branch/.style={
        rectangle,
        rounded corners=7pt,
        draw=none,
        fill=teal!14,
        blur shadow={shadow blur steps=5, shadow xshift=1.2pt, shadow yshift=-1.2pt},
        align=center,
        minimum width=3.25cm,
        minimum height=0.9cm,
        font=\small\bfseries
    },
    metric/.style={
        rectangle,
        rounded corners=6pt,
        draw=none,
        fill=orange!14,
        blur shadow={shadow blur steps=5, shadow xshift=1.2pt, shadow yshift=-1.2pt},
        align=center,
        minimum width=3.2cm,
        minimum height=1.15cm,
        font=\scriptsize
    },
    final/.style={
        rectangle,
        rounded corners=7pt,
        draw=none,
        fill=purple!12,
        blur shadow={shadow blur steps=5, shadow xshift=1.2pt, shadow yshift=-1.2pt},
        align=center,
        minimum width=4.0cm,
        minimum height=0.9cm,
        font=\small\bfseries
    },
    arrow/.style={-Latex, thick, draw=black!55}
]

\node[input] (real) {Full OD Data};
\node[input, right=of real] (sample) {1\% Training Sample};
\node[input, right=of sample] (gen) {Generated Population};

\node[main, below=1.15cm of sample] (eval) {Evaluation Framework};

\node[branch, below left=1.15cm and 1.55cm of eval] (tab) {Tabular Quality};
\node[branch, below=1.15cm of eval] (seq) {Sequential Quality};
\node[branch, below right=1.15cm and 1.55cm of eval] (cons) {Cross-Stage\\Consistency};

\node[metric, below=0.7cm of tab] (tabm) {
Feasibility\\
Diversity / Diversity$_w$\\
Novelty / Novelty$_w$\\
F1 scores
};

\node[metric, below=0.7cm of seq] (seqm) {
Trip length\\
N-grams\\
Diversity, Novelty, Feasibility
};

\node[metric, below=0.7cm of cons] (consm) {
Mobility status\\
vs.\\
Trip chains
};

\node[final, below=1.65cm of seqm] (final) {Synthetic Population Quality};

\draw[arrow] (real.south) |- (eval.west);
\draw[arrow] (sample) -- (eval);
\draw[arrow] (gen.south) |- (eval.east);

\draw[arrow] (eval) -- (tab);
\draw[arrow] (eval) -- (seq);
\draw[arrow] (eval) -- (cons);

\draw[arrow] (tab) -- (tabm);
\draw[arrow] (seq) -- (seqm);
\draw[arrow] (cons) -- (consm);

\draw[arrow] (tabm.south) |- (final.west);
\draw[arrow] (seqm) -- (final);
\draw[arrow] (consm.south) |- (final.east);

\end{tikzpicture}
\caption{Overview of the evaluation framework for assessing tabular quality, sequential quality, and cross-stage consistency.}
\label{fig:evaluation_framework}
\end{figure}
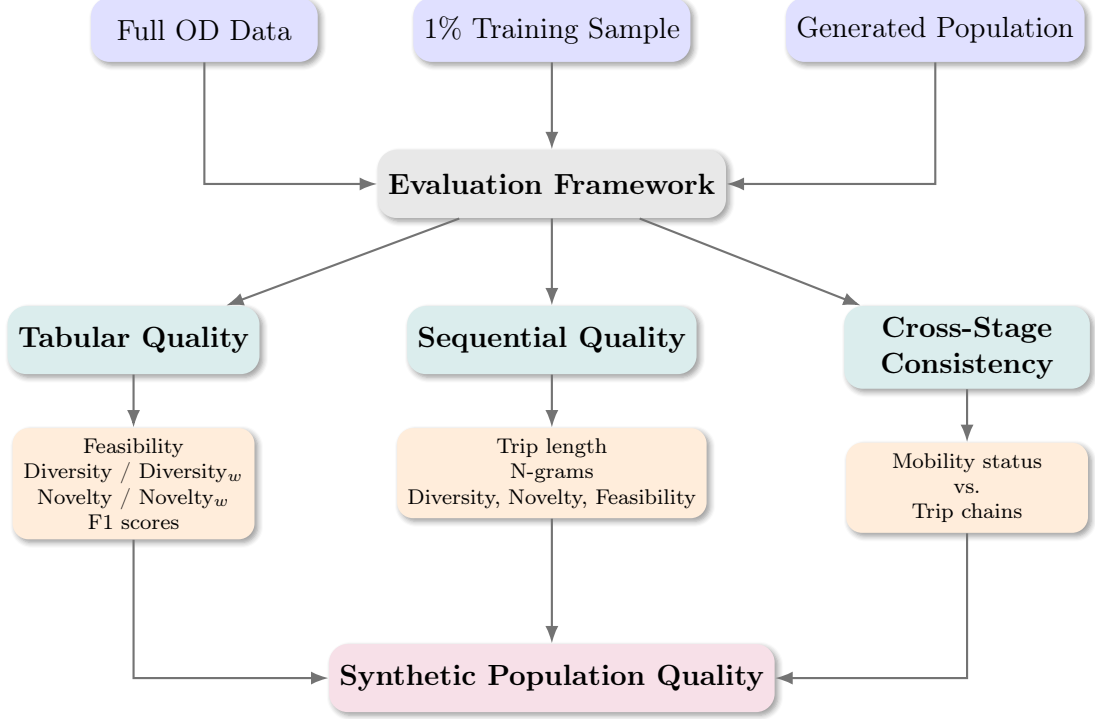

Diversity is evaluated by comparing the number of unique n-gram patterns produced by the generated data with those observed in the real data and the sample. This indicates whether the generated data can reproduce a broad range of sequential patterns rather than only repeating the limited sample. Novelty is assessed by identifying generated n-grams that do not appear in the sample dataset. Some of these novel patterns may correspond to valid patterns that exist in the full real dataset but were missed by the small sample. Therefore, we also examine how many missing real patterns are recovered by the generated data. Feasibility is evaluated by identifying generated patterns that are not supported by the real data. These generated-only patterns may indicate unrealistic or invalid sequential combinations. Figure~\ref{fig:evaluation_framework}, illustrates the overall structure of the evaluation framework.

\section{Results and discussion}

The empirical evaluation is conducted using the Montreal OD survey, described in Section~3, which serves as the true population. A 1\% random sample of the OD dataset is designated as training data. This sample captures 628 unique tabular attribute combinations, representing only 14.7\% of the 4,282 unique combinations present in the full dataset. The remaining 3,654 combinations constitute sampling zeros A model that merely memorises the training distribution cannot recover these missing profiles, making their recovery the central challenge addressed by the proposed regularization framework.

Stage~1 is evaluated across four generative model variants: a vanilla WGAN-GP and three regularized extensions that incorporate CLAP, IGP, and LDR. All generated populations are evaluated against the full OD dataset using the feasibility, diversity, novelty, and count-aware metrics defined in Section~4. Stage~2 compares the Transformer and LSTM-Attention models for sequential behavioural attribute generation. Sequential quality is assessed through three complementary metrics. First, the trip-length distribution of the generated population is compared against the full OD dataset to verify that the models reproduce realistic daily travel volumes. Second, to assess whether the models reproduce complete daily activity patterns, we compare the distribution of the most frequent real trip purpose sequences with their corresponding shares in the generated population. Third, the internal structure of the generated trip chains is evaluated by extracting unigrams, bigrams, and trigrams from the departure time, trip purpose, and travel mode sequences, and measuring their feasibility, diversity, and novelty against the real data.

Finally, to assess the coherence between the two stages of the framework, we conduct a validation based on the mobility status attribute (\textit{p\_mobil}). Because mobility status is generated as part of the tabular profile in Stage~1 and directly conditions whether an individual should produce any trips in Stage~2, it serves as a natural bridge between the two stages. Specifically, we examine whether individuals generated as non-mobile in Stage~1 are consistently assigned empty or null trip sequences in Stage~2, and whether mobile individuals receive plausible trip chains. This consistency check provides a direct measure of how well the joint framework preserves the relationship between socio-demographic attributes and sequential travel behaviour.

\subsection{Stage 1: Tabular Attribute Synthesis}

\subsubsection{Distributional Similarity}
\begin{figure}[!htbp]
    \centering
    \includegraphics[width=0.72\textwidth]{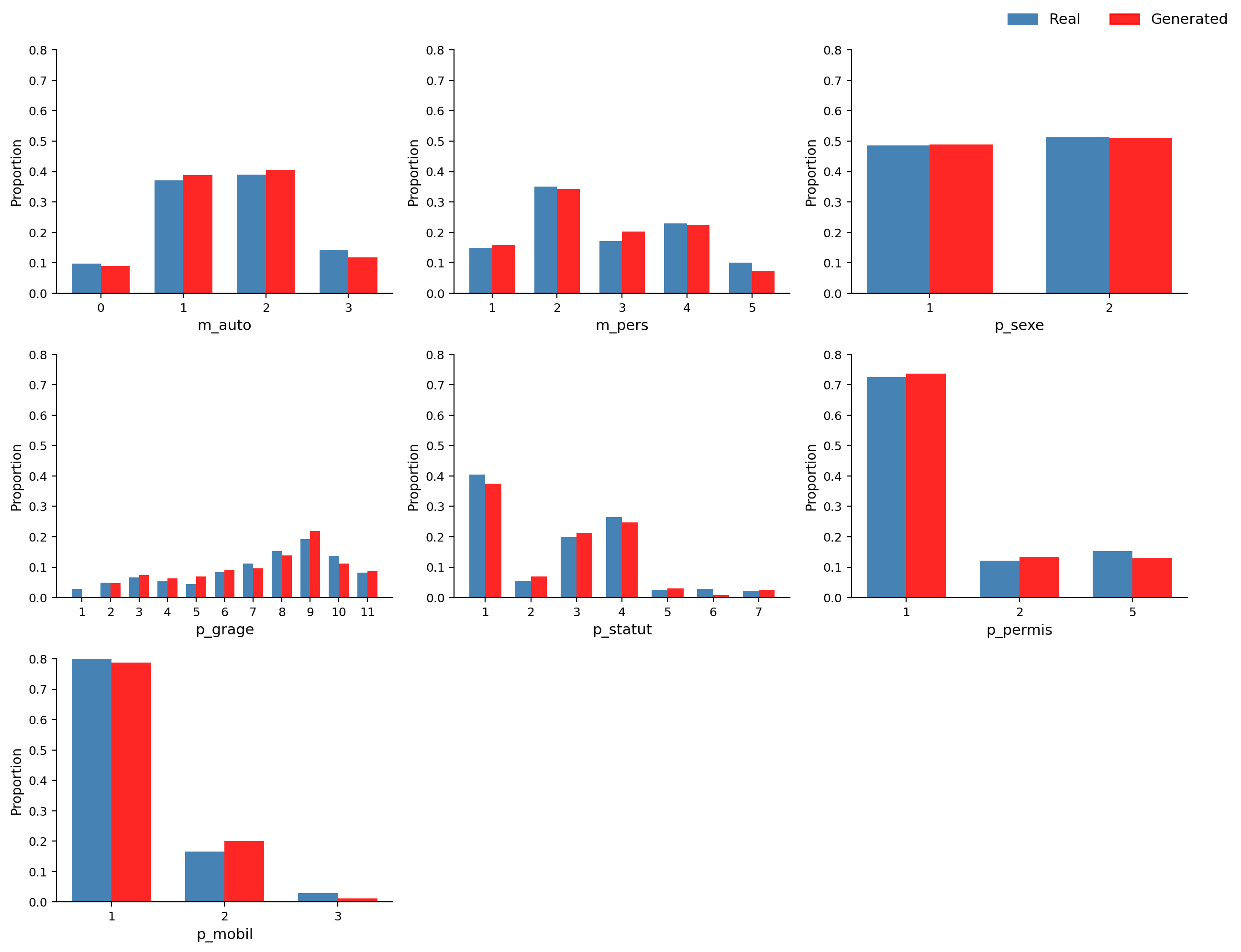}
    \caption{Comparison of marginal probability distributions.}
    \label{fig:marginal_frac001_CLAP_all_features}
\end{figure}

We begin by assessing how well the best-performing model reproduces the marginal attribute distributions of the full OD dataset. Figure~\ref{fig:marginal_frac001_CLAP_all_features} presents side-by-side bar charts comparing the proportions of real (blue) and CLAP-generated (red) individuals across all seven tabular attributes. Across all attributes, the generated data closely follows the real distribution.  There is no clear pattern of over-representing or under-representing any category.  The model also matches simple binary attributes, such as gender and mobility status, very well.  It also captures the uneven distributions of age and employment.  These results show that CLAP can learn the main marginal patterns of the population even when it is trained on only a 1\% sample.

\begin{figure}[!htbp]
    \centering
    \includegraphics[width=0.7\textwidth]{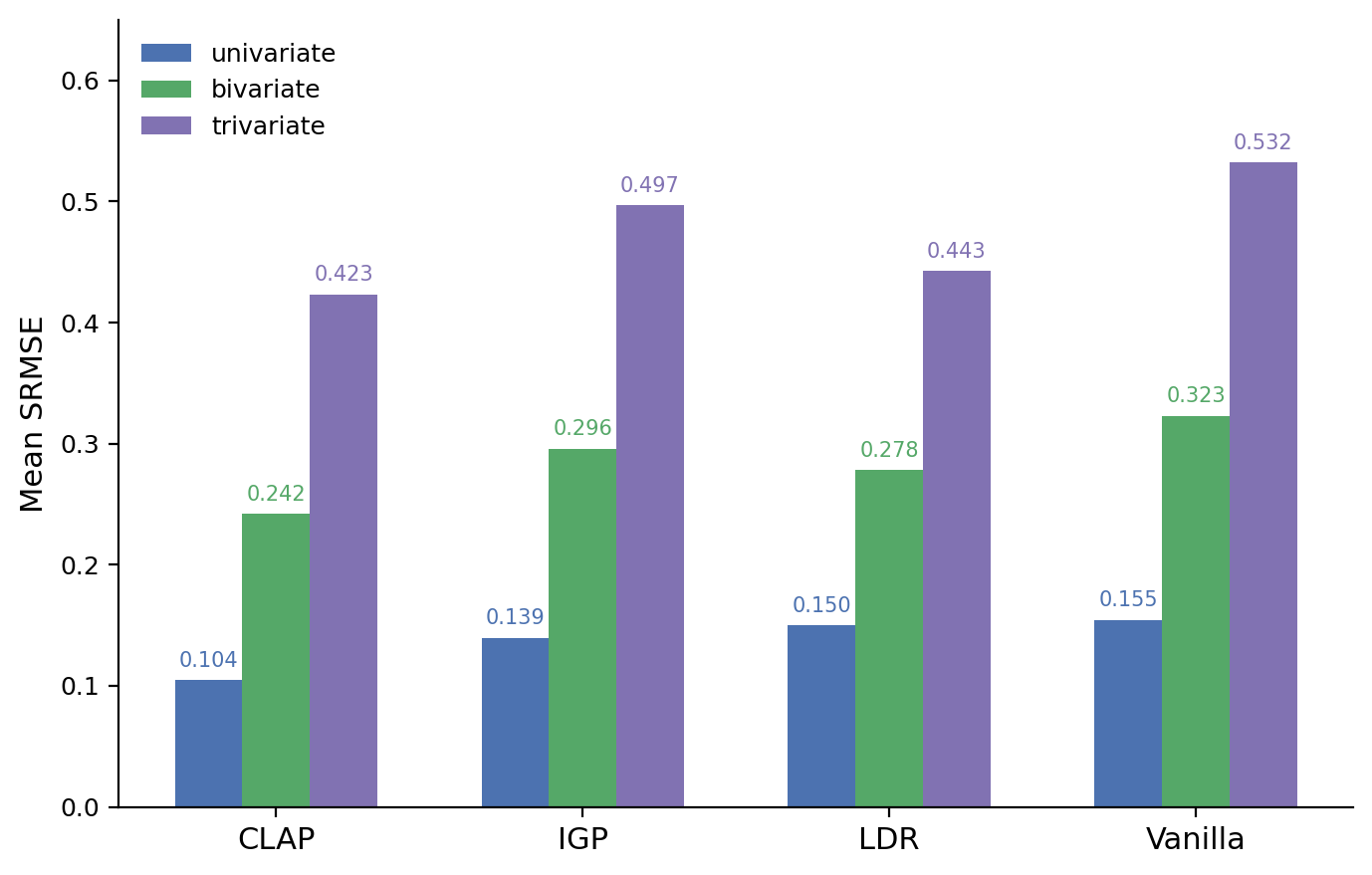}
    \caption{Mean SRMSE across univariate, bivariate, and trivariate distributions for all model variants.}
    \label{fig:srmse_frac001_methods}
\end{figure}

To quantify distributional similarity across model variants and increasing levels of attribute interaction, Figure~\ref{fig:srmse_frac001_methods} reports the mean Standardized Root Mean Square Error (SRMSE) for the vanilla WGAN-GP and the three regularized variants (CLAP, IGP, and LDR) at the univariate, bivariate, and trivariate levels. Overall, the regularized models consistently outperform the vanilla baseline. Among the regularized variants, CLAP achieves the lowest SRMSE across all interaction levels, indicating a stronger ability to reproduce both marginal distributions and higher-order attribute associations.

However, it is important to note that distributional similarity provides only an initial evaluation of synthetic population quality. Although the SRMSE results show that the regularized models, especially CLAP, are better able to match key distributional patterns, further evaluation is required to assess whether the generated data are feasible, novel, diverse, and useful for downstream applications.

\subsubsection{Feasibility}

Table~\ref{tab:combination_metrics} presents the performance of the Vanilla WGAN-GP and the three regularized variants. A first important observation is that all models generate a larger number of unique attribute combinations than those observed in the real dataset. The real population contains 4,282 unique combinations, while the generated datasets contain between 5,412 and 6,408 unique combinations. Even in the most conservative case, the generator produces more combinations than the full real data. This increase in the number of generated combinations is not necessarily a negative result. However, it is important to check whether this extra diversity leads to unrealistic samples.

Feasibility is the first metric used to examine this issue. This metric measures the proportion of generated samples that correspond to valid combinations in the real population. As shown in Table~\ref{tab:combination_metrics}, all models achieve high feasibility. The Vanilla model obtains a feasibility score of 0.898, while the regularized models achieve higher values: 0.935 for CLAP, 0.919 for IGP, and 0.934 for LDR. This means that approximately 90\% to 94\% of the generated individuals represent combinations that genuinely exist in the real data.

The regularized models all outperform the Vanilla model in terms of feasibility. This result shows that adding regularization does not push the generator toward unrealistic attribute profiles. Instead, the regularized models are able to improve the validity of the generated population. The difference between the best and worst feasibility scores is relatively small, around three to four percentage points, which suggests that all models learn the basic structure of the population reasonably well.

The relationship between feasibility and exploration is particularly important. In general, when a model is encouraged to generate more diverse samples, there is a risk that it may also generate more infeasible combinations. However, the results show that the regularized models achieve both higher diversity and higher feasibility than the Vanilla model. This indicates that regularization improves the quality of exploration rather than simply increasing the number of generated combinations. Overall, feasibility is not a major problem for any of the models when training is effective. Most generated samples correspond to valid real-population combinations. 

\subsubsection{Diversity}

As shown in Table~\ref{tab:combination_metrics}, all regularized models achieve higher diversity than the Vanilla model. Vanilla obtains a diversity score of 0.638, while CLAP, IGP, and LDR improve this value to 0.678, 0.741, and 0.714, respectively. Among all models, IGP achieves the highest diversity score, meaning that it recovers 74.1\% of all real combinations. This result shows that IGP encourages the generator to explore a wider part of the attribute space compared with the unregularized baseline.

However, the standard diversity score only checks whether a combination appears or not. It does not consider whether the generated frequency of that combination is close to its true frequency in the real population. For this reason, the weighted diversity metric is more informative. It evaluates not only whether the model recovers real combinations, but also whether it generates them in more realistic proportions.

When weighted diversity is considered, the difference between the regularized models becomes much smaller. IGP and CLAP both achieve a weighted diversity score of 0.684. Vanilla remains the lowest, with a score of 0.666. Although IGP achieves the highest standard diversity score, its advantage largely disappears once population frequencies are taken into account. This suggests that IGP is very effective at recovering a large number of unique combinations, but is less successful at generating them in proportions that match the real population. In contrast, CLAP is able to maintain its performance more consistently when moving from standard diversity to weighted diversity. While its standard diversity score is lower than that of IGP, its weighted diversity remains competitive, indicating that the combinations it recovers are generated in more realistic proportions. 

\subsubsection{Novelty}

Novelty focuses on an even more challenging task. It measures how many valid combinations that were absent from the 1\% training sample are recovered by the generated data. In this experiment, there are 3,654 unseen combinations. These combinations exist in the full real population but are not present in the small training sample. Therefore, a high novelty score indicates that the model is not only memorizing the training sample, but is also able to recover valid profiles beyond what it directly observed during training.

The novelty results show that all regularized models outperform the Vanilla model. Vanilla achieves a novelty score of 0.600, while CLAP, IGP, and LDR improve this value to 0.637, 0.706, and 0.676, respectively. IGP again achieves the highest score, recovering more than 70\% of the unseen valid combinations. This confirms that IGP has the strongest ability to explore beyond the limited 1\% sample and recover combinations that were missing from the training data.

The weighted novelty results provide a stricter evaluation. Vanilla has the lowest weighted novelty score, with a value of 0.479. The regularized models perform better, with CLAP reaching 0.548, LDR reaching 0.545, and IGP achieving the highest value of 0.579. These results show that the regularized models are not only better at finding unseen valid combinations, but also better at generating them in more realistic frequencies. However, the drop from standard novelty to weighted novelty also shows that recovering unseen combinations is easier than reproducing their correct population proportions.

These results highlight the importance of using both standard and weighted metrics. Standard diversity and novelty are useful for measuring coverage, but they can make a model appear stronger if it generates many combinations without matching their real frequencies. Weighted diversity and weighted novelty provide a stricter and more realistic evaluation because they consider both coverage and proportional accuracy. Therefore, the weighted metrics are especially important for judging the quality of the synthetic population.

\begin{table}[ht]
\centering
\caption{Combination-level validity, diversity, and novelty results for the generated datasets.}
\label{tab:combination_metrics}
\resizebox{\textwidth}{!}{%
\begin{tabular}{l c c c c c c c c c}
\hline
Method &
$n_{\text{real}}$ &
$n_{\text{sample}}$ &
$n_{\text{unseen}}$ &
$n_{\text{fake}}$ &
Feasibility &
Diversity &
\textbf{Div. (w)} &
\textbf{Novelty} &
\textbf{Nov. (w)} \\
\hline
Vanilla & \multirow{4}{*}{4282} & \multirow{4}{*}{628} & \multirow{4}{*}{3654} & 6351 & 0.898 & 0.638 & 0.666 & 0.600 & 0.479 \\
CLAP    &                       &                      &                       & 5412 & 0.935 & 0.678 & 0.684 & 0.637 & 0.548 \\
IGP     &                       &                      &                       & 6408 & 0.919 & 0.741 & 0.684 & 0.706 & 0.579 \\
LDR     &                       &                      &                       & 5460 & 0.934 & 0.714 & 0.690 & 0.676 & 0.545 \\
\hline
\end{tabular}%
}
\end{table}

\subsubsection{Comparison of Regularization Terms}

The final F1 scores provide a combined evaluation of feasibility and coverage. Since the weighted metrics are stricter and more informative than their standard versions, the weighted F1 scores, especially $F1_{D_w}$ and $F1_{N_w}$, are the primary basis for comparison in this section. As shown in Table~\ref{tab:f1_metrics}, all regularized models achieve higher F1 scores than the Vanilla model in both diversity and novelty evaluations.

An important question is which F1 score should be used as the main criterion for comparing the models. One possible option is $F1_D$, which combines feasibility  with standard diversity. A stricter comparison can be made using the weighted F1 scores. Among these, $F1_{N_w}$ is the most important metric for this study. This is because novelty is more closely related to the sampling-zero problem than diversity. Diversity evaluates coverage over all real combinations, including those that may already be present in the 1\% training sample. In contrast, novelty focuses specifically on valid real combinations that were absent from the training sample. Therefore, novelty directly measures whether the model can recover sampling-zero combinations rather than simply reproduce what it has already seen. For this reason, the most reliable validation should make the evaluation stricter in two ways. First, it should use the weighted version of the metric. Second, it should focus on novelty rather than diversity.

Based on $F1_{N_w}$, the ranking of the models is IGP, CLAP, LDR, and Vanilla. IGP achieves the highest score of 0.711, showing that it provides the best trade-off between generating valid combinations and recovering unseen combinations in realistic proportions. For completeness, all F1 variants are reported in Table~\ref{tab:f1_metrics}, but the weighted novelty F1 score is treated as the main reference for selecting the best model.

\begin{table}[ht]
\centering
\caption{F1-based evaluation combining feasibility with diversity and novelty metrics.}
\label{tab:f1_metrics}
\begin{tabular}{l c c c c}
\hline
Method &
$F1_D$ &
{\boldmath{$F1_{D_w}$}} &
{\boldmath{$F1_N$}} &
{\boldmath{$F1_{N_w}$}} \\
\hline
Vanilla & 0.746 & 0.765 & 0.719 & 0.625 \\
CLAP    & 0.786 & 0.790 & 0.758 & 0.691 \\
IGP     & 0.820 & 0.784 & 0.799 & 0.711 \\
LDR     & 0.809 & 0.793 & 0.784 & 0.688 \\
\hline
\end{tabular}
\end{table}

\subsection{Stage 2 — Sequence Attribute Synthesis}

\subsubsection{Trip Length and Activity Pattern}
\begin{figure}[H]
    \centering
    \includegraphics[width=0.72\textwidth]{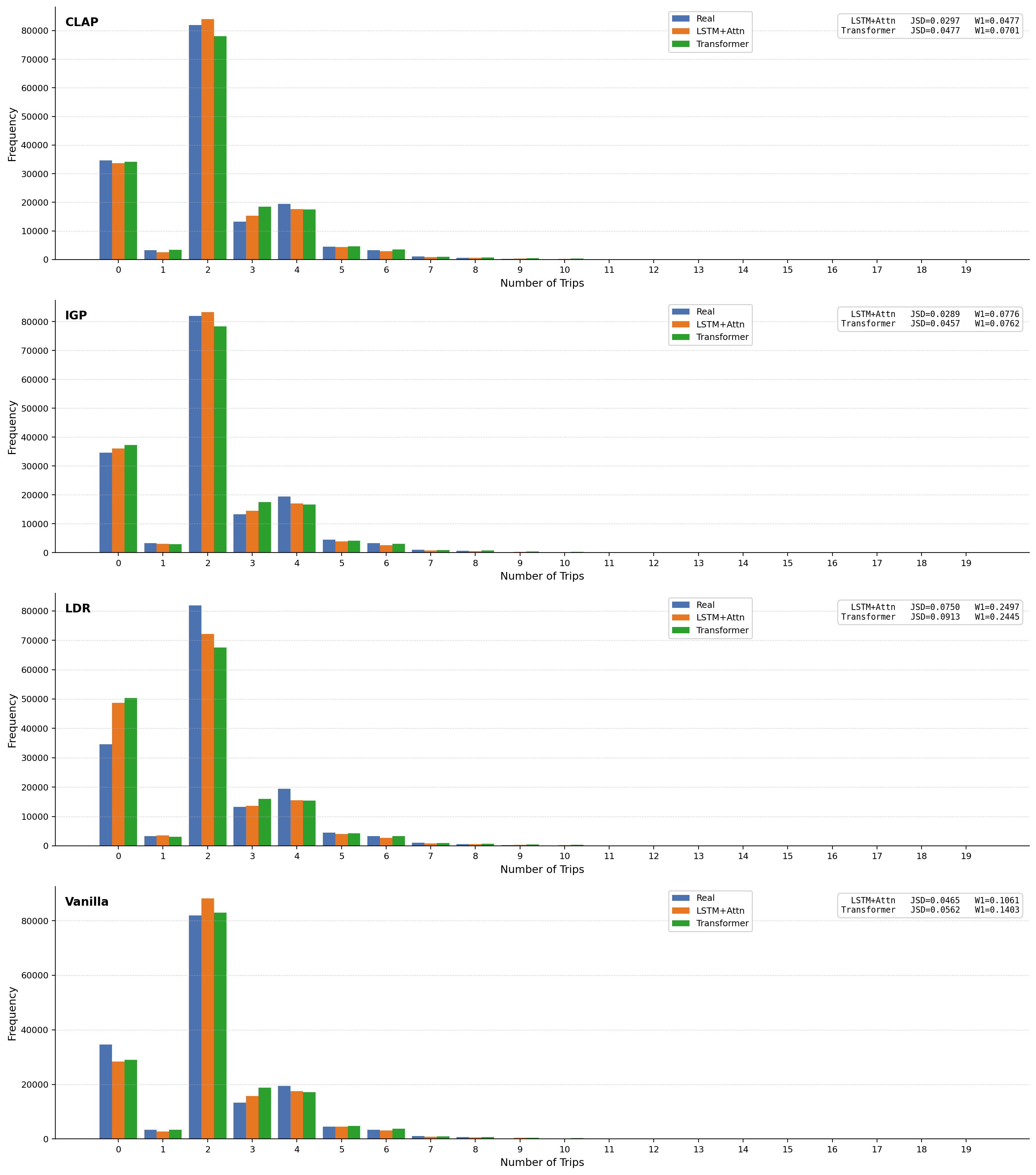}
    \caption{Trip-length distribution comparison between the real OD data and the generated outputs.}
    \label{fig:trip_length}
\end{figure}

The trip-length distribution shows how many trips each individual makes per day. If the generated population behaves similarly to the real population, the generated trip length distribution should be close to the real one. Figure~\ref{fig:trip_length} compares the real OD distribution with the outputs of the LSTM-Attention and Transformer sequence models for each of the four tabular models: CLAP, IGP, LDR, and Vanilla. The real distribution is shown in blue, the LSTM-Attention output in orange, and the Transformer output in green. Two metrics are used to measure distributional similarity: Jensen-Shannon Divergence (JSD) and Wasserstein-1 distance (W1). For both metrics, lower values indicate a closer match to the real distribution.

The first comparison is between the two sequence models. As shown in Figure~\ref{fig:trip_length}, LSTM-Attention gives lower JSD values than the Transformer for the regularized tabular models. This means that LSTM-Attention matches the general shape of the real trip-length distribution more closely. However, the W1 values are very close between the two sequence models, and in some cases the Transformer obtains a slightly lower W1. This suggests that although LSTM-Attention better captures the overall probability pattern, the two models are quite similar in terms of how far the generated trip counts are from the real ones.

The second comparison is the effect of the Stage 1 tabular model. The choice of tabular model clearly affects how well the sequence model reproduces the trip-length distribution. Among the regularized models, CLAP and IGP produce the strongest results. In particular, CLAP combined with LSTM-Attention achieves the best W1 score, showing the closest match to the overall shape of the real distribution. Compared with the Vanilla baseline, CLAP provides a clear improvement, with an approximate W1 difference of 0.058. This indicates that a better tabular generation stage can also improve the quality of the sequence-generation stage.

\begin{figure}[H]
\centering
\includegraphics[width=0.85\textwidth]{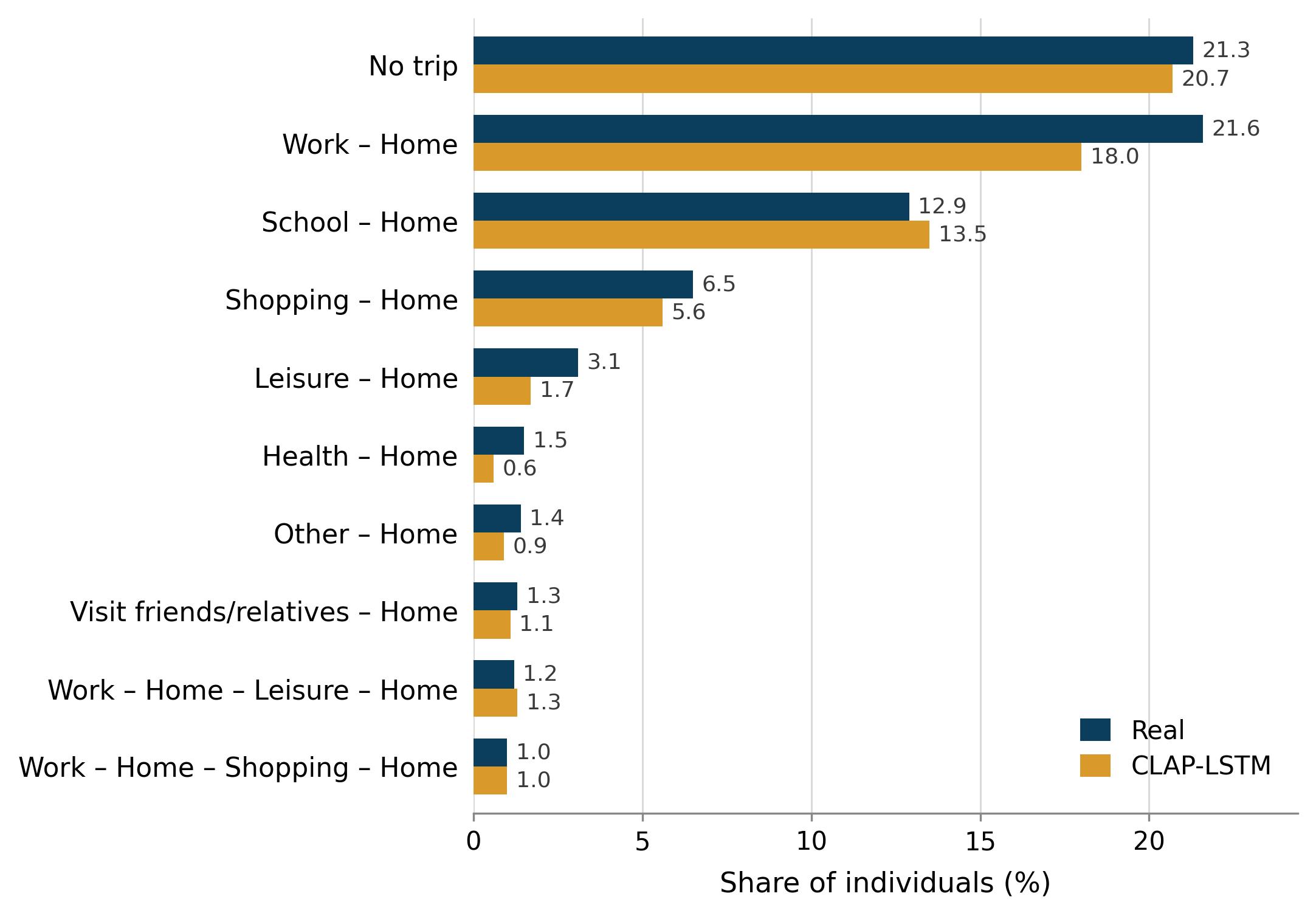}
\caption{Distribution of the ten most frequent real activity sequences and their corresponding shares in the CLAP-LSTM generated population.}
\label{fig:top_activity_sequences}
\end{figure}

To further assess whether the generated data reproduce complete and interpretable activity chain patterns, we compare the distribution of the ten most frequent real activity sequences with their corresponding shares in the CLAP-LSTM generated population, as shown in Figure~\ref{fig:top_activity_sequences}.

These dominant sequences account for a substantial share of the observed activity patterns and therefore provide a useful benchmark for evaluating whether the model captures the main structure of daily travel behaviour. The real dataset contains 162,588 records and 6,389 unique activity sequences, while the CLAP-LSTM generated population contains the same number of records and 6,058 unique sequences. This indicates that the generated population preserves a comparable level of sequence diversity. As shown in Figure~\ref{fig:top_activity_sequences}, the generated shares closely follow the real distribution across the major activity-sequence patterns.

\subsubsection{N-gram Analysis of Trip Chains}

This section evaluates the sequential quality of the generated trip chains using n-gram analysis. Three levels of sequential structure are considered. As the n-gram level increases, the evaluation becomes stricter because the model must reproduce more complex sequential dependencies. For each feature and n-gram level, the generated sequences are compared with the real OD data using diversity, novelty, and feasibility.

\begin{figure}[H]
    \centering
    \includegraphics[width=0.9\textwidth]{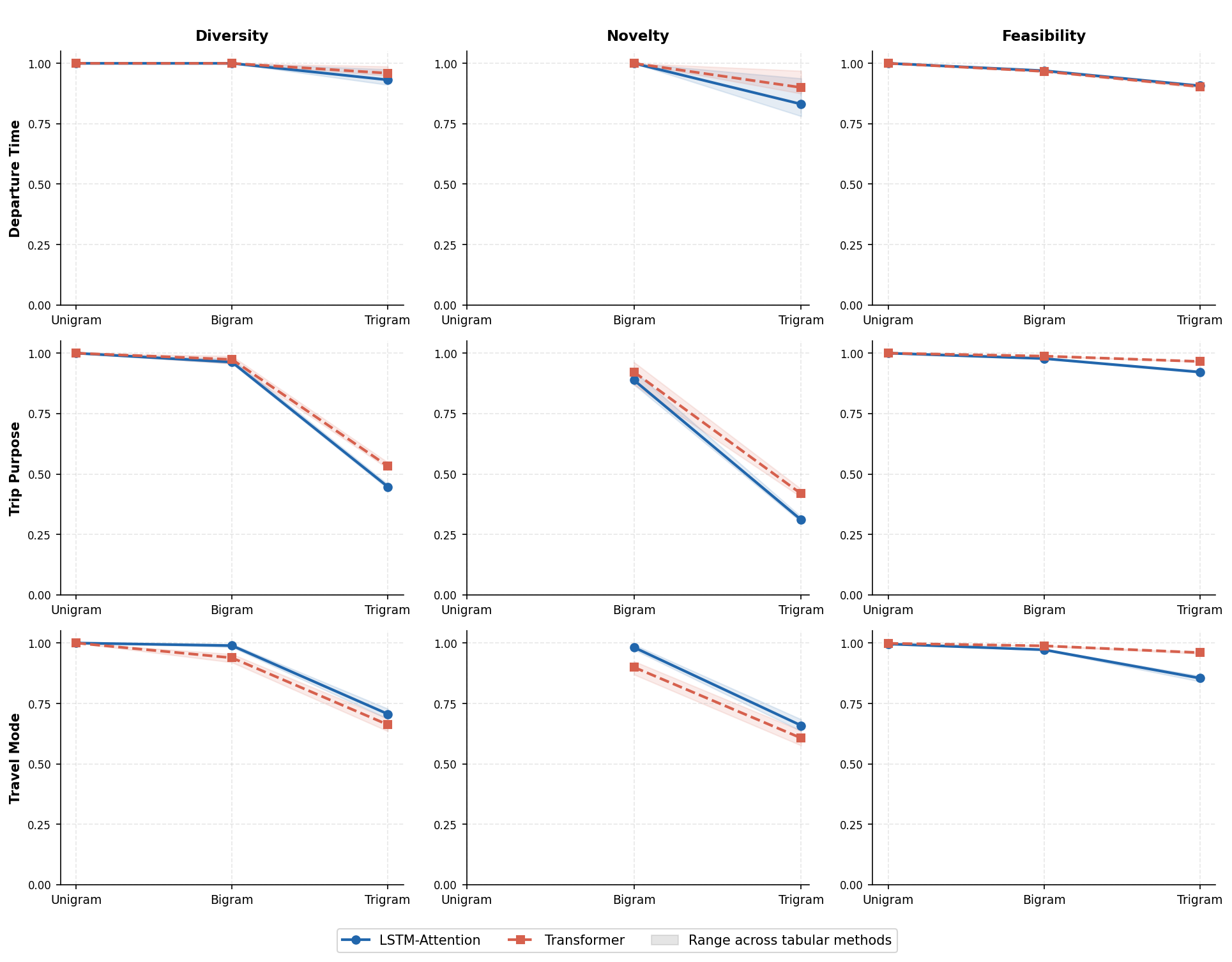}
    \caption{N-gram evaluation of LSTM-Attention and Transformer. The shaded area shows the range of results across the Stage 1 tabular model variants.}
    \label{fig:ngram_lstm_transformer}
\end{figure}

Figure~\ref{fig:ngram_lstm_transformer} compares the n-gram performance of LSTM-Attention and Transformer across the three sequence features: departure time, trip purpose, and travel mode. Overall, both models perform well at the unigram and bigram levels, but their behaviour becomes more different at the trigram level, where the task is more difficult.

In terms of diversity and novelty, the Transformer also performs well for departure time and trip purpose. For these two features, it generally matches or improves over LSTM-Attention, especially at the trigram level. This indicates that the Transformer is effective not only at generating valid sequences, but also at recovering a wider range of real and unseen sequential patterns. However, the pattern is different for travel mode. In this feature, LSTM-Attention performs better than the Transformer in diversity and novelty, showing that it recovers more travel-mode combinations, although with lower feasibility.

The decrease in performance at the trigram level is expected. Trigrams represent longer and more specific sequential patterns, so they are harder to reproduce than unigrams or bigrams. This is especially clear when comparing the three features. Departure time has the best trigram performance in diversity and novelty because it has a smaller number of possible combinations, with only 79 trigram patterns. In contrast, travel mode has 875 trigram combinations, and trip purpose has 1,322 combinations. As the number of possible combinations increases, it becomes more difficult for the model to recover all valid patterns, especially unseen ones. Therefore, the lower diversity and novelty scores for trip purpose and travel mode are mainly due to the larger and more complex combinatorial space.

One important observation is that the Transformer generally maintains higher feasibility than LSTM-Attention, especially at the trigram level. This difference is most visible for travel mode. Based on Table~\ref{tab:combined_ngram_results}, the feasibility of LSTM-Attention drops to around 0.84 for travel-mode trigrams, while the Transformer keeps feasibility close to 0.96. A similar pattern can be observed for trip purpose, where the LSTM-Attention feasibility is around 0.92, while the Transformer reaches approximately 0.97. This suggests that the Transformer is better at keeping longer generated sequences valid, even when the sequential patterns become more complex. The detailed numerical results for all features, n-gram levels, and model combinations are provided in Table~\ref{tab:combined_ngram_results}.

\FloatBarrier
\begin{table}[H]
\centering
\scriptsize
\setlength{\tabcolsep}{3pt}
\renewcommand{\arraystretch}{0.92}
\caption{Combined N-gram evaluation results for all features using a 1\% sample. Results are shown separately for each feature and model family.}
\label{tab:combined_ngram_results}
\resizebox{\textwidth}{!}{%
\begin{tabular}{lllcccccc}
\toprule
\multirow{2}{*}{Feature} & \multirow{2}{*}{Method} & \multirow{2}{*}{N-gram}
& \multicolumn{3}{c}{LSTM}
& \multicolumn{3}{c}{Transformer} \\
\cmidrule(lr){4-6} \cmidrule(lr){7-9}
& & & Diversity & Novelty & Feasibility & Diversity & Novelty & Feasibility \\
\midrule

\multirow{12}{*}{\texttt{d\_grhre\_vec}}
& \multirow{3}{*}{vanilla}
& unigram & 1.000 & --    & 1.000 & 1.000 & --    & 1.000 \\
& & bigram  & 1.000 & 1.000 & 0.969 & 1.000 & 1.000 & 0.968 \\
& & trigram & 0.911 & 0.781 & 0.907 & 0.949 & 0.875 & 0.906 \\

& \multirow{3}{*}{IGP}
& unigram & 1.000 & --    & 1.000 & 1.000 & --    & 1.000 \\
& & bigram  & 1.000 & 1.000 & 0.969 & 1.000 & 1.000 & 0.965 \\
& & trigram & 0.911 & 0.781 & 0.906 & 0.949 & 0.875 & 0.899 \\

& \multirow{3}{*}{LDR}
& unigram & 1.000 & --    & 1.000 & 1.000 & --    & 1.000 \\
& & bigram  & 1.000 & 1.000 & 0.969 & 1.000 & 1.000 & 0.965 \\
& & trigram & 0.924 & 0.813 & 0.909 & 0.962 & 0.906 & 0.902 \\

& \multirow{3}{*}{CLAP}
& unigram & 1.000 & --    & 1.000 & 1.000 & --    & 1.000 \\
& & bigram  & 1.000 & 1.000 & 0.969 & 1.000 & 1.000 & 0.967 \\
& & trigram & 0.975 & 0.938 & 0.904 & 0.949 & 0.875 & 0.903 \\
\midrule

\multirow{12}{*}{\texttt{d\_motif\_vec}}
& \multirow{3}{*}{vanilla}
& unigram & 1.000 & --    & 1.000 & 1.000 & --    & 1.000 \\
& & bigram  & 0.957 & 0.870 & 0.977 & 0.957 & 0.870 & 0.988 \\
& & trigram & 0.439 & 0.301 & 0.921 & 0.530 & 0.416 & 0.966 \\

& \multirow{3}{*}{IGP}
& unigram & 1.000 & --    & 1.000 & 1.000 & --    & 1.000 \\
& & bigram  & 0.963 & 0.889 & 0.977 & 0.988 & 0.963 & 0.987 \\
& & trigram & 0.458 & 0.323 & 0.918 & 0.533 & 0.417 & 0.964 \\

& \multirow{3}{*}{LDR}
& unigram & 1.000 & --    & 1.000 & 1.000 & --    & 1.000 \\
& & bigram  & 0.975 & 0.926 & 0.978 & 0.969 & 0.907 & 0.987 \\
& & trigram & 0.445 & 0.309 & 0.924 & 0.523 & 0.404 & 0.966 \\

& \multirow{3}{*}{CLAP}
& unigram & 1.000 & --    & 1.000 & 1.000 & --    & 1.000 \\
& & bigram  & 0.957 & 0.870 & 0.977 & 0.975 & 0.926 & 0.989 \\
& & trigram & 0.449 & 0.314 & 0.921 & 0.548 & 0.438 & 0.966 \\
\midrule

\multirow{12}{*}{\texttt{d\_mode\_main}}
& \multirow{3}{*}{vanilla}
& unigram & 1.000 & --    & 0.996 & 1.000 & --    & 0.999 \\
& & bigram  & 0.996 & 0.993 & 0.972 & 0.933 & 0.890 & 0.989 \\
& & trigram & 0.696 & 0.648 & 0.857 & 0.648 & 0.589 & 0.962 \\

& \multirow{3}{*}{IGP}
& unigram & 1.000 & --    & 0.996 & 1.000 & --    & 0.998 \\
& & bigram  & 0.992 & 0.986 & 0.970 & 0.937 & 0.897 & 0.986 \\
& & trigram & 0.710 & 0.663 & 0.841 & 0.672 & 0.619 & 0.955 \\

& \multirow{3}{*}{LDR}
& unigram & 1.000 & --    & 0.996 & 1.000 & --    & 0.999 \\
& & bigram  & 0.983 & 0.973 & 0.972 & 0.920 & 0.870 & 0.988 \\
& & trigram & 0.686 & 0.636 & 0.862 & 0.635 & 0.576 & 0.962 \\

& \multirow{3}{*}{CLAP}
& unigram & 1.000 & --    & 0.996 & 1.000 & --    & 0.999 \\
& & bigram  & 0.983 & 0.973 & 0.971 & 0.954 & 0.925 & 0.989 \\
& & trigram & 0.728 & 0.684 & 0.849 & 0.688 & 0.637 & 0.960 \\
\bottomrule
\end{tabular}%
}
\end{table}
\FloatBarrier

To identify the best overall model combination, the n-gram results are further summarized using an F1 score. For unigrams, the F1 score is computed using feasibility and diversity, because unigrams do not have a novelty value. For bigrams and trigrams, the F1 score is computed using feasibility and novelty, because these are more directly related to the recovery of unseen sequential patterns.

\begin{figure}[H]
    \centering
    \includegraphics[width=0.95\textwidth]{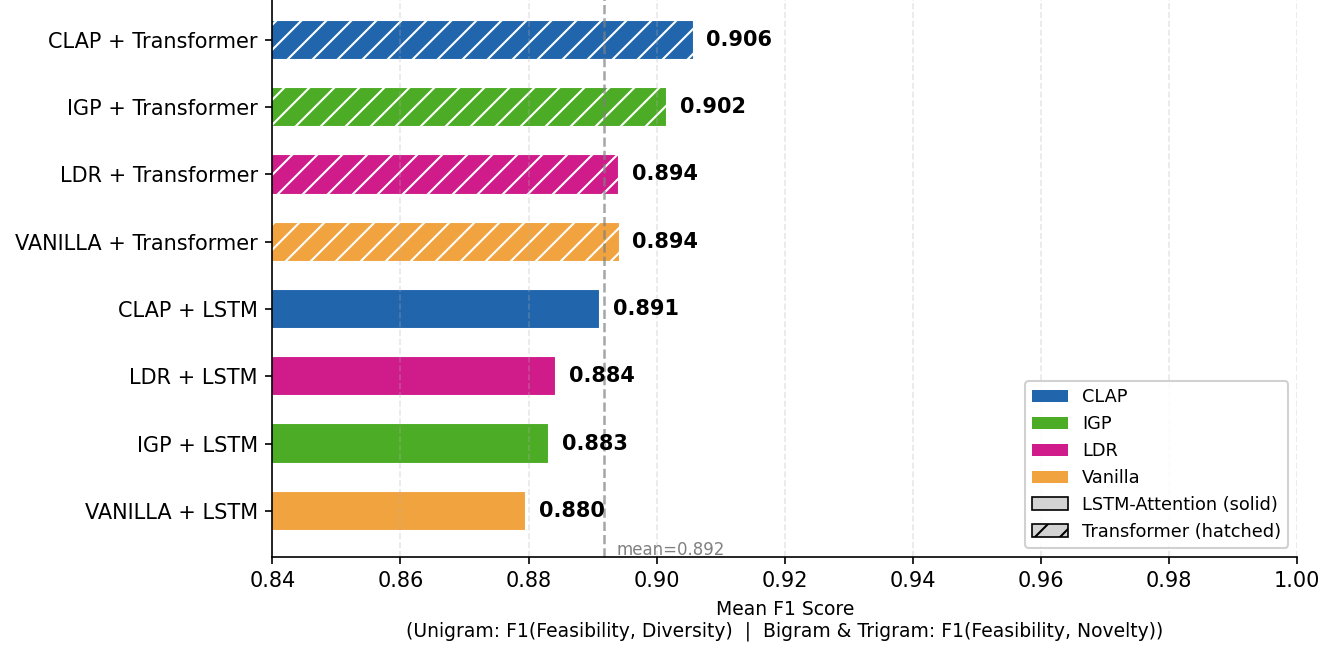}
    \caption{Overall sequential quality ranking of the eight model combinations based on the mean F1 score.}
    \label{fig:ngram_f1_ranking}
\end{figure}

Figure~\ref{fig:ngram_f1_ranking} reports the mean F1 score for all eight model combinations. The final score is averaged across the three sequence features, departure time, trip purpose, and travel mode, and across the three n-gram levels. This provides a single overall measure of sequential quality.

The ranking shows that Transformer-based combinations achieve the highest overall scores. The best-performing model is CLAP + Transformer, with a mean F1 score of 0.906. This confirms that the Transformer provides a consistent improvement in overall sequential quality.

The figure also shows that the Stage 1 tabular model still affects the final sequence quality. CLAP gives the best result for both sequence architectures. Compared with the Vanilla baseline, the regularized tabular models generally improve the final F1 score. 

\subsection{Cross-Stage Consistency: Mobility Validation}

To assess whether the two stages of the framework are coherent with each other, the mobility status attribute, $p_{\text{mobil}}$, is used as a bridge between the tabular and sequential outputs. In the real OD dataset, individuals with $p_{\text{mobil}}=2$ are non-mobile and should not have any trips. Therefore, all three sequential attributes, departure time, trip purpose, and travel mode, should be empty sequences represented as $[0]$. In contrast, individuals with $p_{\text{mobil}}=1$ are mobile and should have at least one trip. Based on this rule, two types of violations are counted: non-mobile individuals who are incorrectly assigned trips, and mobile individuals who are incorrectly assigned no trips.

The purpose of this analysis is not to compare or rank the models, but to check whether the generated tabular and sequential components remain semantically consistent after the two-stage generation process. As shown in Table~\ref{tab:pmobil_consistency}, all model combinations achieve very high consistency, with accuracy values above 98\% for LSTM-Attention and above 99\% for Transformer. This indicates that the generated mobility status is largely consistent with the generated trip sequences.

These results confirm that the two-stage framework preserves meaningful semantic coherence between the tabular and sequential outputs. In other words, the sequence-generation stage does not behave independently of the tabular attributes; instead, it generally produces trip chains that are logically compatible with the mobility status assigned in the first stage.

\begin{table*}[ht]
\centering
\caption{Consistency check between \texttt{p\_mobil} and the three trip-related features.}
\label{tab:pmobil_consistency}
\resizebox{\textwidth}{!}{
\begin{tabular}{lcccccccc}
\toprule
\multirow{2}{*}{Method} 
& \multicolumn{4}{c}{LSTM} 
& \multicolumn{4}{c}{Transformer} \\
\cmidrule(lr){2-5} \cmidrule(lr){6-9}
& \makecell{Violations\\all 3 = [0]} 
& \makecell{Violations\\all 3 $\neq$ [0]} 
& \makecell{Total\\violations} 
& \makecell{Accuracy\\(\%)} 
& \makecell{Violations\\all 3 = [0]} 
& \makecell{Violations\\all 3 $\neq$ [0]} 
& \makecell{Total\\violations} 
& \makecell{Accuracy\\(\%)} \\
\midrule

IGP     & 2278 & 16 & 2294 & 98.59 & 771 & 2 & 773 & 99.52 \\
LDR     & 2804 & 9  & 2813 & 98.27 & 647 & 5 & 652 & 99.60 \\
vanilla & 1343 & 22 & 1365 & 99.13 & 466 & 8 & 474 & 99.71 \\
CLAP    & 1089 & 17 & 1106 & 99.32 & 316 & 5 & 321 & 99.80 \\

\bottomrule
\end{tabular}
}
\vspace{0.5em}

\footnotesize{
Real data baseline: 0 violations and 100\% consistency accuracy. 
Total number of samples: 162,588.
}
\end{table*}

\section{Conclusions}

This paper proposed a two-stage framework for generating synthetic populations with both tabular socio-demographic attributes and sequential travel-behaviour attributes. The framework addresses three key limitations in existing population synthesis methods: achieving feasibility, diversity, and novelty in tabular attribute synthesis, integrating the modelling of sequential behavioural attributes such as trip purpose, departure time, and travel mode, and improving the evaluation of synthetic population quality beyond conventional metrics. Together, these challenges are addressed through a framework that recovers valid but unseen combinations, avoids structurally invalid profiles, generates behaviorally realistic trip chains, and evaluates synthetic populations using feasibility, diversity, novelty, and count-aware measures.

In the first stage, a WGAN-GP model was extended with regularization terms designed to improve the recovery of sampling-zero combinations while reducing the generation of structurally invalid profiles. The results show that regularization improves the quality of tabular synthesis compared with the vanilla WGAN-GP baseline. All regularized models achieved higher feasibility, diversity, novelty, and count-aware scores than the unregularized model. Among the proposed regularization methods, IGP produced the strongest overall performance when evaluated using the count-aware novelty F1 score, indicating that it provides the best balance between generating feasible individuals and recovering valid combinations absent from the 1\% training sample. 

In the second stage, Transformer and LSTM-Attention models were used to generate sequential behavioural attributes conditioned on the synthesized tabular profiles. The results show that both models are able to reproduce key characteristics of daily trip chains, including trip-length distributions, activity patterns, and n-gram patterns in departure time, trip purpose, and travel mode sequences. While LSTM-Attention performs competitively in matching some trip-length distributions, the Transformer generally provides stronger sequential validity, especially for higher-order trip-chain patterns. 

The cross-stage mobility validation further confirms the coherence of the proposed framework. Generated mobility status from the tabular stage was highly consistent with the generated trip sequences in the sequential stage, with consistency rates above 98\% for LSTM-Attention and above 99\% for Transformer-based models. This result indicates that the sequential generator does not operate independently of the tabular attributes, but instead produces behavioural outputs that remain logically compatible with individual socio-demographic profiles.

While the proposed regularization terms improve the balance between novelty, diversity, and feasibility, they remain soft training mechanisms rather than hard constraints. Therefore, they cannot fully prevent structural violations or guarantee the recovery of all sampling zero combinations. Their performance also depends on hyperparameter choices and the quality of the limited training sample.

Several directions remain for future work. First, the framework should be tested on additional regions and travel survey datasets to assess its transferability and robustness across different urban contexts. Second, future research should investigate household-level constraints and interactions, since many travel decisions are shaped by relationships among household members. Third, the sequence-generation stage could be extended to include richer spatial information, such as activity locations or origin-destination patterns. Fourth, the framework could be extended from a static synthesis approach to a dynamic population synthesis model by incorporating demographic changes over time, such as ageing, household formation, employment transitions, vehicle ownership changes, and shifts in mobility status. Finally, future work should evaluate the generated synthetic populations within full activity-based modelling pipelines to measure how improvements in population synthesis affect downstream travel-demand forecasts.

\section*{Acknowledgments}
This study is funded by the Canada First Research Excellence Fund under the Bridging Divides program.

\section*{Declaration of generative AI}

During the preparation of this work, the author(s) used ChatGPT to assist with language editing, clarity improvement, manuscript formatting, and the preparation of tables and figures. After using this tool, the author(s) reviewed and edited the content as needed and take full responsibility for the content of the published article.

\section*{Data availability}

The data that support the findings of this study are from the 2018 Montreal Origin-Destination survey. Restrictions apply to the availability of these data, which were used under license for the current study and are not publicly available from the authors.

\bibliographystyle{elsarticle-harv}
\bibliography{references}
\end{document}